# Road Curb Extraction from Mobile LiDAR Point Clouds

Sheng Xu, Ruisheng Wang, Han Zheng[1]

**Abstract:** Automatic extraction of road curbs from uneven, unorganized, noisy and massive 3D point clouds is a challenging task. Existing methods often project 3D point clouds onto 2D planes to extract curbs. However, the projection causes loss of 3D information which degrades the performance of the detection. This paper presents a robust, accurate and efficient method to extract road curbs from 3D mobile LiDAR point clouds. Our method consists of two steps: 1) extracting the candidate points of curbs based on the proposed novel energy function and 2) refining the candidate points using the proposed least cost path model. We evaluated our method on a large-scale of residential area (16.7GB, 300 million points) and an urban area (1.07GB, 20 million points) mobile LiDAR point clouds. Results indicate that the proposed method is superior to the state-of-the-art methods in terms of robustness, accuracy and efficiency. The proposed curb extraction method achieved a completeness of 78.62% and a correctness of 83.29%. These experiments demonstrate that the proposed method is a promising solution to extract road curbs from mobile LiDAR point clouds.

## 1. Introduction

Mobile LiDAR System (MLS) is a newly emerging technology which collects 3D information of objects while vehicles drive at a posted speed [1]. It becomes more and more popular in analyzing 3D point clouds because of its high density, efficiency and cost-effectiveness and provides the possibility to extract the micro-objects such as road curbs.

Road curb extraction from 3D point clouds is a basis for several types of research, such as road surface analysis, driving simulation, safe parking, autonomous driving and traffic environment understanding. However, point clouds acquired by MLS are often found to be uneven, unorganized, noisy and massive, thereby making the curb detection a challenging task.

In this paper, we present a robust, accurate and efficient method to extract road curbs from mobile LiDAR point clouds. The main contributions of our work are the following: 1) We propose a novel energy function to extract candidate points of curbs from mobile LiDAR point clouds. 2) We propose a least cost path model to link candidate points into complete curbs. 3) We conduct a comprehensive evaluation of the proposed method using a large-scale data.

The paper is organized as follows. In Section 2, we review the state-of-the-art methods related to road curb extraction. In Section 3, we propose a new energy function to extract the candidate points of curbs and a least cost path model to connect the candidate points into the global optimization curbs. In Section 4, we evaluate the robustness, accuracy and complexity of our method. The conclusions are presented in Section 5.

## 2. Related work

A straightforward method for road curb detection usually makes use of elevation information. For example, algorithms [2-6] focus on detecting objects in terms of elevation difference. It is possible to obtain road curbs by elevation filtering, however at the compromise of robustness. There is no reliable cues to design adaptive thresholds for the low elevation curbs in different scenes. These methods produce attractive results in the straight roads with the same elevation. However they fail to work in occluded, sunken or uphill road areas.

Recent methods [7-10] focus on models that incorporate more prior knowledge, such as width, elevation and density, to form the descriptors for classifying regions and edges. The prior knowledge based method [11] uses a pre-defined curb model in terms of elevation jump, point density and slope change to provide potential location of curbs from mobile LiDAR point clouds. However, this method doesn't work well on datasets with different geometrical features, such as large slopes or uneven road surface, due to the use of non-robust 3D features.

[1]Sheng Xu, Ruisheng Wang, and Han Zheng are with the Department of Geomatics Engineering, Schulich School of Engineering, University of Calgary, Calgary, Alberta, T2N 1N4, Canada. (sheng.xu2@ucalgary.ca; ruiswang@ucalgary.ca; han.zheng@ucalgary.ca)

The typical technique for boundary extraction is the active contour model (Snake) [12]. Snake is widely applied to curb extraction from images or images generated from 3D point clouds' projection [13-15]. However, the projection loses 3D information that will degrade the performance of the extraction. This is also the main drawback of the extraction methods [16-18] based on 2D images. Moreover, Snake needs a manual initialization to start the iteration.

The methods [19-20] combine LiDAR point clouds and the corresponding images to detect road curbs, but fail to work when there are occlusions caused by cars, pedestrians or trees along the road. Moreover, the registration of LiDAR point clouds and images is not reliable due to the duplication of moving objects.

In summary, the challenges in curb detection from mobile LiDAR point clouds are as follows. First, the data is difficult to process due to its uneven, unorganized, noisy and massive nature. Second, there is no much reliable information, such as color, intensity and texture, for segmentation or classification. Third, since the LiDAR sensors mounted on the moving vehicle scan objects line by line, there will be misalignments or duplicates of moving objects and occlusions.

This paper aims to provide a robust solution for road curb extraction from mobile LiDAR point clouds. In a pre-processing step, we remove the non-ground points by using elevation histogram and organize the ground point clouds into voxels. Then, we extract the candidate points of curbs using the proposed energy function. Last, we use the proposed least cost path model to complete optimal curbs.

Comparison with the existing curb extraction algorithms, our method has no risk of losing 3D information. Since all existing methods lack experiments on a large-scale data, we evaluate our algorithm on a large-scale residential and a medium sized urban data to verify the proposed method.

## 3. The method

### 3.1. Definition of 3D sampling density and density gradient

In 2D images, the gradient shows the increase or decrease in the magnitude of the intensity. However, to the best of our knowledge, no unified definition of the gradient exists for 3D point clouds to date. This paper analogizes the gradient in 2D to obtain the gradient definition in 3D which only uses the geometric information of the MLS measurement.

In this paper, the gradient concept is extended to 3D point clouds through considering the points' density in a local area. At first, voxels are generated for the point cloud. Then, the intensity of each voxel is defined by the points' density, i.e. the number of the point inside the voxel. Finally, our 3D sampling density gradient is calculated by the difference of the intensity between adjacent voxels. The intensity is approximated by the number of points in a local area.

One existing 3D gradient definition is based on elevation difference between adjacent points. In Fig.1 (a), the elevation along Z axis is increasing evenly, so the gradient is a constant along Z axis and zero along Y axis. However, the gradient along X axis varies because of different elevations. This is not desirable, because the gradients along the normal direction of the façade (i.e. along X axis) are different as shown in Fig.1 (a). Our definition of 3D sampling density gradient is based on the intensity difference between the adjacent points. Our gradient is zero along either Y axis or Z axis and a constant along X axis as shown in Fig.1 (b), which better represents the real situation. We use this new 3D intensity and sampling density gradient definition throughout this paper.

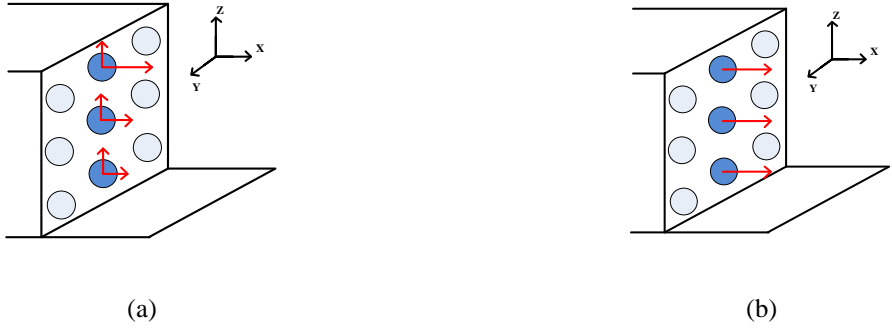

(a)                                        (b)

Fig.1. The different definitions of the 3D gradient. (a) Elevation difference. (b) Intensity difference.

Similar to the change produced by a shift for a pixel in the 2D image [21], we use Eq.(1) to define the magnitude of change (squared) in the diagonal direction for a point $p(x, y, z)$ in a small shift $(\Delta x, \Delta y, \Delta z)$,

$$C(x,y,z) = \sum_{\Delta x, \Delta y, \Delta z} \left[ I_{\Delta x+x, \Delta y+y, \Delta z+z} - I_{x,y,z} \right]^2 \qquad (1)$$

$I_{x,y,z}$ is the intensity which is the number of points in a local area around $p$. Our point clouds are uneven and unorganized. The Euclidean distance between each point is various and neighbors of each point are unknown. Thus, we use voxel of a suitable size to represent each local area and organize 3D point clouds in a sparse 3D matrix. The value of each voxel is the intensity, which is the number of points in each voxel. Finally, we can deal our data like pixels in 2D image with the voxel based representation. The $(\Delta x, \Delta y, \Delta z)$ means the coordinate difference between two voxels which can be also treated as a directional vector, for example (1,0,0) is the direction of $X$ axis, (1,1,0) is the direction of 45 degrees in the $XOY$, (1,0,1) is the direction of 45 degrees in the $XOZ$, etc. The coordinate difference should be an integer.

### 3.2. Classification of the road areas

There are mainly three regions in road point clouds as shown in Fig.2: roadway, sidewalk and curb. The curb connects the roadway and the sidewalk and is usually lower than 0.25 meters. The Euclidean distance between two points in our data is larger than 0.004 meters. The elevation difference between sidewalk and roadway is small. Thus, the road curb detection cannot heavily rely on the elevation difference.

Suppose that the point clouds are aligned with a 3D coordinate system $O$-$XYZ$ as shown in Fig.2, the curb is in parallel with $X$ axis and $XOZ$ plane, roadway and sidewalk are in parallel with $XOY$ plane.

Denote $G_x$, $G_y$ and $G_z$ as our sampling density gradients of a voxel along $X$ axis, $Y$ axis and $Z$ axis directions respectively as shown in Fig.3. There will be three primary situations for our gradients of a voxel in Fig.2.

(1) The voxel within the surface: there is only one large gradient, such as the large $G_y$ in the curb areas and the large $G_z$ in the roadway or sidewalk areas.

(2) The voxel in the intersection of two surfaces: there are more than one large gradient, such as the large $G_y$ and $G_z$ along the curb edges.

(3) The voxel in the intersection of three mutually non-parallel surfaces: all gradients are large, such as the curb corners.

Therefore, we conclude that if gradients of a voxel are large in more than one direction, the points in this voxel potentially belong to the curb.

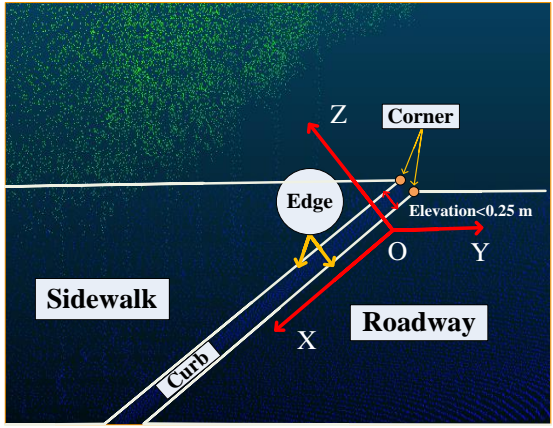 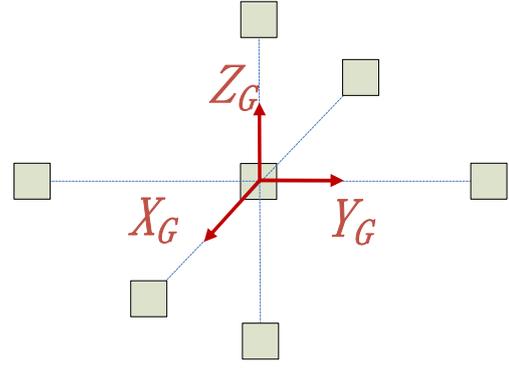

Fig.2. Three areas in road point clouds, namely sidewalk, roadway and curb.

Fig.3. Sampling density gradients of a voxel in each axis direction.

### 3.3. Mathematical model

Eq.(1) can be rewritten as Eq.(2), the Taylor series expansion with $O(\Delta x^2, \Delta y^2, \Delta z^2)$ as the remainder term.

$$C(x,y,z) = \sum_{\Delta x, \Delta y, \Delta z} \left[ \Delta x \cdot G_x + \Delta y \cdot G_y + \Delta z \cdot G_z + O(\Delta x^2, \Delta y^2, \Delta z^2) \right]^2 \qquad (2)$$

$G_x$, $G_y$ and $G_z$ are the sampling density gradients which are defined as

$$G_x = \left(\frac{\partial I}{\partial x}\right) \approx I \cdot Cube_x$$
$$G_y = \left(\frac{\partial I}{\partial y}\right) \approx I \cdot Cube_y \qquad (3)$$
$$G_z = \left(\frac{\partial I}{\partial z}\right) \approx I \cdot Cube_z$$

$Cube_x$, $Cube_y$ and $Cube_z$ are three 3×3×3 operators extended from Sobel [22] as shown in Fig.4. $I$ is a 3×3×3 matrix whose elements are the intensity of each voxel and its neighbors.

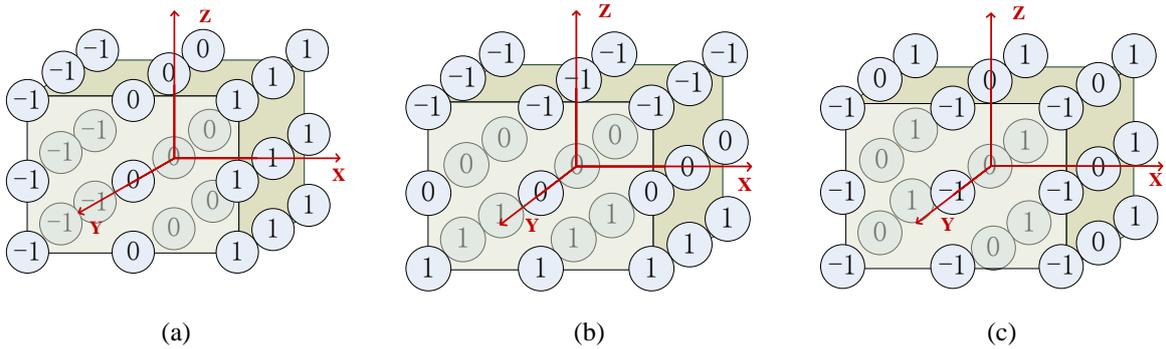

Fig.4. 3D Sobel. (a) $Cube_x$. (b) $Cube_y$. (c) $Cube_z$.

We obtain $C(x,y,z)$ for each voxel as

$$\begin{aligned} C(x,y,z) &= (G_x \cdot \Delta x)^2 + (G_y \cdot \Delta y)^2 + (G_z \cdot \Delta z)^2 \\ &+ 2 \times \Delta x \cdot \Delta y \cdot G_x \cdot G_y + 2 \times \Delta y \cdot \Delta z \cdot G_y \cdot G_z + 2 \times \Delta x \cdot \Delta z \cdot G_x \cdot G_z \\ &= (\Delta x, \Delta y, \Delta z) \mathbf{M} (\Delta x, \Delta y, \Delta z)^T \geq 0 \end{aligned} \qquad (4)$$

where **M** is

$$\begin{pmatrix} G_x^{\,2} & G_x \cdot G_y & G_x \cdot G_z \\ G_x \cdot G_y & G_y^{\,2} & G_y \cdot G_z \\ G_x \cdot G_z & G_y \cdot G_z & G_z^{\,2} \end{pmatrix}$$

**M** is a semi-positive symmetric matrix. Its eigenvectors are mutually orthogonal and eigenvalues $\alpha$, $\beta$ and $\gamma$ are not less than 0. According to Section 3.2, three possibilities will be observed for each voxel: (1) the voxel belongs to the surface area, when only one large eigenvalue exists; (2) the voxel belongs to two surfaces' intersection when two large eigenvalues are observed; and (3) the voxel belongs to the intersection area of three mutually non-parallel surfaces, when three large eigenvalues are observed.

In 2D images, the edges have a large gradient in one direction. The intersection of two edges is the corners which have a large gradient in more than one direction. Similarly, for our sampling density gradient in 3D point clouds, a large gradient in one direction means planar surface areas. The edges are the intersection of two surfaces. Thus, the candidate points of curb edges have large gradients in at least two directions. Now our sampling density gradients are related to the eigenvalues of the matrix **M**. We show different ranges of eigenvalues in Fig.5. There are four areas, ① all of eigenvalues are small, ② only one of eigenvalues is large, ③ two of eigenvalues are large and ④ all of eigenvalues are large.

We can decide the area of each voxel by thresholding method. However, it is difficult to tune different thresholds for each eigenvalue. Thus, we form an energy function based on the eigenvalues $\alpha$, $\beta$ and $\gamma$. The energy corresponding to each voxel can map to the areas in Fig.5. If the result of mapping is in ③ or ④, the voxel is chosen as the candidate curb edges.

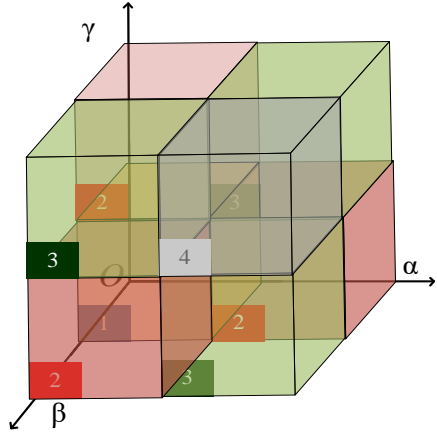

Fig.5. Four areas for different ranges of the eigenvalue $\alpha$, $\beta$ and $\gamma$. ① All eigenvalues are small. ② Only one large eigenvalue. ③ Two large eigenvalues. ④ All eigenvalues are large.

### 3.4. Construction and analysis of the energy function

We design our energy function based on a triangle $Q$. The energy will be related to the area $S$ of this triangle. The eigenvalues $\alpha$, $\beta$ and $\gamma$ are related to angles $\alpha'$, $\beta'$ and $\gamma'$ of $Q$. The side $a$ is in opposite to $\alpha'$, $b$ is in opposite to $\beta'$ and $c$ is in opposite to $\gamma'$ as shown in Fig.6. As described in Section 3.3, sampling density gradients large in two or three directions are now related to the triangle $Q$ with at least two large angles.

Fig.6 (a) is the initial triangle. Increase the height of the triangle $Q$ to change the base angles $\alpha'$ and $\beta'$ together. One can find that the area $S$ is growing as increasing the two base angles. If the angle $\gamma'$ is infinitely close to 0, $S$ achieves the maximum as shown in Fig.6 (b).

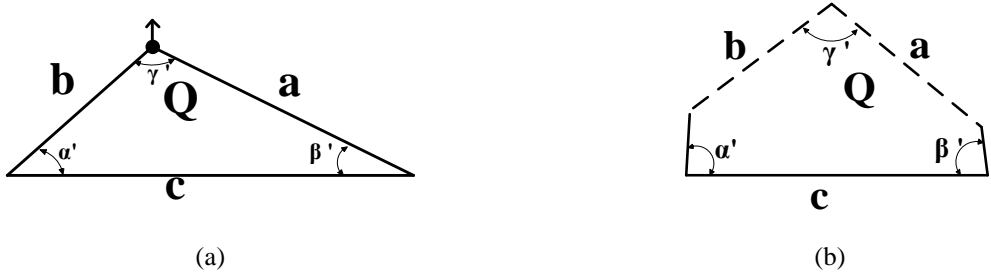

(a)                                (b)

Fig.6. Related triangle. (a) Initial triangle $Q$. (b) Increase the height of $Q$.

Our idea is to relate the energy $E$ to the area $S$ to ensure that, if $E$ is large, there will be more than one large eigenvalue. The challenges are relations among angles, eigenvalues and the length of sides. The following is the derivation of the energy function.

Let

$$\varphi = \frac{\pi}{\arctan\alpha + \arctan\beta + \arctan\gamma} \qquad (5)$$

$$\alpha' = \arctan\alpha \times \varphi, \quad \gamma' = \arctan\gamma \times \varphi, \quad \beta' = \arctan\beta \times \varphi$$

The area $S$ is derived according to the sine theorem:

$$\begin{aligned} S &= \frac{1}{2} a \times b \sin\gamma' = \frac{1}{2} \times \frac{\sin\beta'}{\sin\gamma'} \times \frac{\sin\alpha'}{\sin\gamma'} \times \sin\gamma' \times c^2 = \frac{c^2}{2} \times \frac{\sin\alpha' \times \sin\beta'}{\sin(\pi - (\alpha' + \beta'))} \\ &= \frac{c^2}{2} \times \frac{\sin\alpha' \times \sin\beta'}{\sin\alpha'\cos\beta' + \cos\alpha'\sin\beta'} = \frac{c^2}{2} \times \frac{\tan\alpha' \times \tan\beta'}{\tan\alpha' + \tan\beta'} \end{aligned} \qquad (6)$$

and

$$\begin{aligned} \tan\alpha' &= \tan(\arctan\alpha \times T) = \tan(\frac{\arctan\alpha}{\arctan\alpha + \arctan\beta + \arctan\gamma} \times \pi) \\ \tan\beta' &= \tan(\arctan\beta \times T) = \tan(\frac{\arctan\beta}{\arctan\alpha + \arctan\beta + \arctan\gamma} \times \pi) \end{aligned} \qquad (7)$$

Both *tan* and *arctan* are monotonically increasing functions, so we have

$$\begin{aligned} S &\propto \tan\alpha' \propto \arctan\alpha \propto \alpha \\ S &\propto \tan\beta' \propto \arctan\beta \propto \beta \end{aligned} \qquad (8)$$

Relating the energy $E$ to the area $S$ we have

$$E \propto \frac{\alpha\beta}{\alpha + \beta} \times c^2 \qquad (9)$$

The side $c$ can be regarded as the weighting coefficient. To visualize the energy $E$, we calculate the result of Eq.(9) using a range [0, 10000] for $\alpha$ and $\beta$. We can define $c$ as a constant 1 or a variable. The sum of the eigenvalue $\alpha$ and $\beta$ is a real number, thus we can define $c$ as $(\alpha+\beta)$. Under this definition, the larger the $\alpha$ and $\beta$, the larger the $E$. Fig.7 (a) and Fig.7 (b) are the visualization of $E$ when $c$ is a constant 1 and a variable $(\alpha+\beta)$, respectively. As seen from Fig.7, when $c$ is defined as $(\alpha+\beta)$, large energy corresponds to the situation where both $\alpha$ and $\beta$ are large.

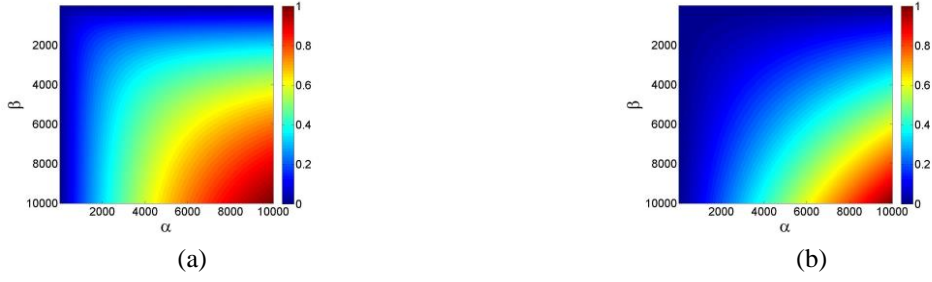

(a)                                               (b)

Fig.7 Visualization of the energy $E$. (a) $c=1$. (b) $c=(\alpha+\beta)$.

With increasing the height of the triangle $Q$ by enlarging only one base angle $\alpha'$ as shown in Fig.8, the area $S$ is also increased. It means that large energy $E$ can be observed in non-curb areas with only one large eigenvalue. To address this problem, we refined the Eq.(9).

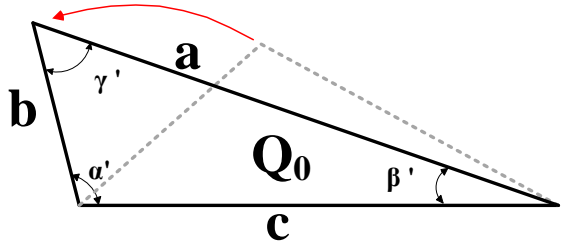

Fig.8. Increase the area $S$ by enlarging only one base angle $\alpha'$.

In the obtuse triangle $Q_0$, the angle $\beta'$ and $\gamma'$ are small. If we treat the side $a$ as the fixed bottom side and do the same analysis as in $Q$, its area $S$ will be smaller than the triangle with a large $\beta'$ and $\gamma'$. Thus, we calculate the sum of areas based on all three sides (Eq.(10)) to ensure that the sum $S'$ is large only when there is more than one large angle and consequently we have

$$S' = \frac{c^2}{2} \times \frac{\tan\alpha' \times \tan\beta'}{\tan\alpha' + \tan\beta'} + \frac{a^2}{2} \times \frac{\tan\beta' \times \tan\gamma'}{\tan\beta' + \tan\gamma'} + \frac{b^2}{2} \times \frac{\tan\alpha' \times \tan\gamma'}{\tan\alpha' + \tan\gamma'} \qquad (10)$$

For our energy function in 3D space, $c$ is unified as $(\alpha+\beta+\gamma)$ and we relate our energy to the sum $S'$ as

$$E \propto \left[ \frac{\alpha\beta}{\alpha+\beta} \times (\alpha+\beta+\gamma)^2 + \frac{\alpha\gamma}{\alpha+\gamma} \times (\alpha+\beta+\gamma)^2 + \frac{\gamma\beta}{\gamma+\beta} \times (\alpha+\beta+\gamma)^2 \right] \qquad (11)$$

$(\alpha+\beta+\gamma)^2$ is regarded as the coefficient and we let

$$E = \left( \frac{\alpha\beta}{\alpha+\beta} + \frac{\alpha\gamma}{\alpha+\gamma} + \frac{\gamma\beta}{\gamma+\beta} \right) \times (\alpha+\beta+\gamma)^2 \qquad (12)$$

We plot the energy $E$ based on different ranges of the eigenvalue $\alpha$, $\beta$ and $\gamma$ using Eq.(12) as shown in Fig.9. The magnitude of $E$ is scaled to [0, 255] and $\alpha$, $\beta$ and $\gamma$ are ranged from 1 to 100. From Fig.9 (a), any two of $\alpha$, $\beta$ and $\gamma$ are large results in a large $E$ and the largest $\alpha$, $\beta$ and $\gamma$ causes the maximum $E$ indicated by the red corner. If the voxel is in the block or surface, $E$ is small (<40) as shown in Fig.9 (b) corresponds to areas ①  and ② in Fig.5. If the voxel is in the curb edges or corners, $E$ is large (>157) as shown in Fig.9 (c) corresponds to areas ③ and ④ in Fig.5. The other $E$ ([40, 157]) is the unreliable areas correspond to area ⑤ in Fig.5. This energy function meets goals mentioned in Section 3.3 which can be used to detect curbs from road points.

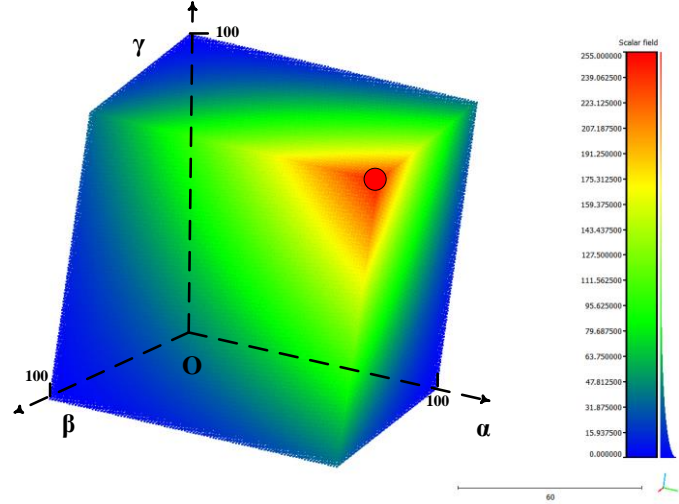

(a)

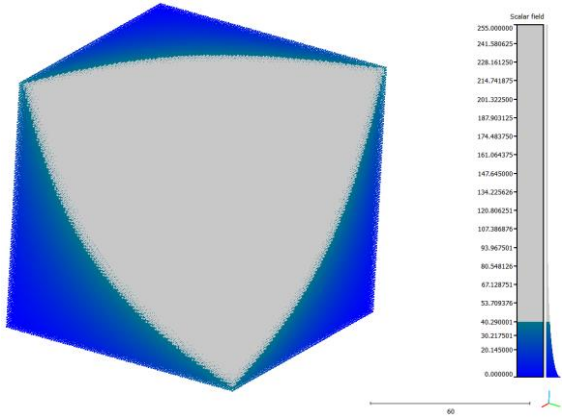

(b)

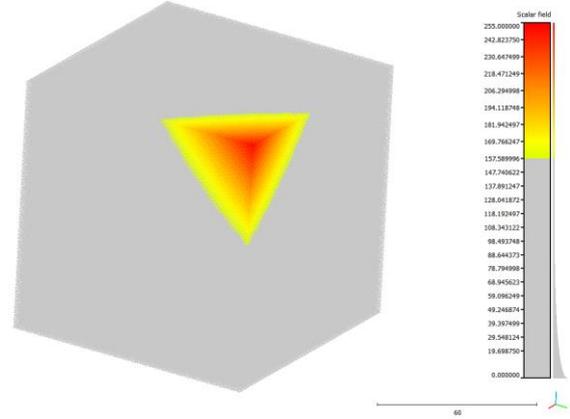

(c)

Fig. 9. The 3D visualization of the energy function $E$. (a) Plot of $E$ based on different ranges of $\alpha$, $\beta$ and $\gamma$. (b) $E$ is small (<40) which corresponds to areas ①and ②in Fig.5. (c) $E$ is large (>157) which corresponds to areas ③ and ④ in Fig.5.

In this paper, voxels corresponding to the top 20% energy are chosen as candidate curbs. Practically, we obtain the energy $E$ without the computation of eigenvalues or eigenvectors of the matrix $\mathbf{M}$. To calculate $E$ in a low complexity, we decompose the matrix $\mathbf{M}$ as $\mathbf{M_{xoy}}$, $\mathbf{M_{xoz}}$ and $\mathbf{M_{yoz}}$ on the *XOY*, *YOZ* and *XOZ* plane, respectively, as follows:

$$\mathbf{M_{xoy}} = \begin{pmatrix} X_G \cdot X_G & X_G \cdot Y_G \\ X_G \cdot Y_G & Y_G \cdot Y_G \end{pmatrix}, \mathbf{M_{xoz}} = \begin{pmatrix} X_G \cdot X_G & X_G \cdot Z_G \\ X_G \cdot Z_G & Z_G \cdot Z_G \end{pmatrix}, \mathbf{M_{yoz}} = \begin{pmatrix} Y_G \cdot Y_G & Y_G \cdot Z_G \\ Y_G \cdot Z_G & Z_G \cdot Z_G \end{pmatrix}$$

Now Eq.(12) can be calculated effectively by

$$E = \frac{Det(\mathbf{M_{xoy}})}{Tr(\mathbf{M_{xoy}})} \times Tr(\mathbf{M})^2 + \frac{Det(\mathbf{M_{xoz}})}{Tr(\mathbf{M_{xoz}})} \times Tr(\mathbf{M})^2 + \frac{Det(\mathbf{M_{yoz}})}{Tr(\mathbf{M_{yoz}})} \times Tr(\mathbf{M})^2 \qquad (13)$$

where *Det* means the determinant and *Tr* is the trace. From Eq.(13), there is no need to compute the eigenvalues or eigenvectors of the matrix. The complexity of the energy computation is linear time which is significant for large-scale point cloud processing.

It is worth pointing out that $Det(M_{xoy})$, $Det(M_{xoz})$ and $Det(M_{yoz})$ are 0, because we use the discrete form Eq.(3) to obtain the sampling density gradient approximately. To overcome this problem, we smooth the input $I$ by convolving a 3×3×3 Gaussian kernel $h$

$$h(i,j,k) = \frac{1}{\sqrt{2\pi\sigma^2}} e^{-\frac{(i-1)^2+(j-1)^2+(k-1)^2}{2\sigma^2}} \quad (14)$$

where $\sigma$ is the standard deviation of the voxels in the convolutional operation. For example, the voxel at $(x, y, z)$ is smoothed as

$$I_h(x,y,z) = I(x,y,z) \otimes h = \sum_{i=0}^{2}\sum_{j=0}^{2}\sum_{k=0}^{2}(I(x,y,z) \cdot h(i,j,k)) \quad (15)$$

where the output $I_h$ is the smoothed input $I$. From the associative property of convolution for linear system, the gradient of the convolution is equal to the convolution of the gradient as shown in Eq.(16). Thus, the gradient calculated from $I_h$ is also smoothed and $Det(M_{xoy})$, $Det(M_{xoz})$ and $Det(M_{yoz})$ are not 0.

$$\frac{\partial(I(x,y,z) \otimes h)}{\partial x} = \frac{\partial I(x,y,z)}{\partial x} \otimes h \quad (16)$$

### 3.5. Least cost path model

Candidate curb edges obtained by optimizing the energy function, are incomplete and noisy. We can refine the curbs by line fitting methods, such as Least square fitting (LS) [23], Hough transform (HT) [24] and RANSAC [25]. Nevertheless, these methods highly rely on the number of candidate curb points and don't consider the non-candidate points and the cost for linking candidate points. The linking of the curb candidate points with few inliers is problematic. We propose a new robust method to link curb candidate points into a complete curb.

Our model consists of a data term and a smoothness term as shown in Eq.(17) to represent different refinement paths. $N$ is the number of nodes in the final path, $(u, v, w)$ is the coordinate of the current node $i$ and $j$ is the node prior to $i$ in the path. $L_N$ is the cost of the refinement path.

$$L_N = \sum_{i}^{N}\left(Data_i(u,v,w) + Smoothness_{i,j}(u,v,w)\right) \quad (17)$$

The data term refers to the cost of a path containing all selected nodes and the smoothness term is the cost connecting node $i$ with the prior node $j$ in the path. The definitions of data term and smoothness term are shown in Eq.(18).

$$Data_i(u,v,w) = \begin{cases} 0, & if\ i \in CanPs \\ penaltyD, & if\ i \notin NCanPs \\ penaltyV, & if\ i \in VirPs \end{cases} \quad (18)$$

$$Smoothness_{i,j}(u,v,w) = penaltyS \times Dis(i,j),$$

where $CanPs$ is the set of voxels filled with candidate points, $NCanPs$ is the set of voxels filled with non-candidate points and $VirPs$ is the set of virtual nodes for occlusions.

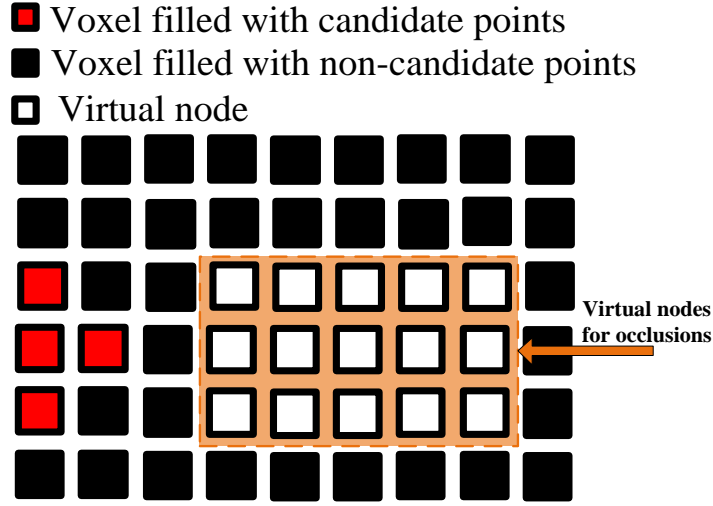

Fig.10. Nodes formed by voxels.

As shown in Fig.10, each voxel forms a node to construct LCPM. If the node $i$ is a voxel filled with candidate points, the data term is 0; if the node $i$ is a voxel filled with non-candidate points, the data term is *penaltyD*; if there is no voxel between two nodes in the axis direction, we fill this no voxel area with virtual nodes and the data term is *peanltyV*.

The smoothness term is calculated by *penaltyS* and the Euclidean distance *Dis* between $i$ and $j$. Each refinement path corresponds to a cost $L_N$ in Eq.(17) and the least cost is the optimal solution.

The graph to be optimized is constructed by the candidate points as described in Section 3.4. Our goal is to connect them into the optimal curbs. Assuming that there are $N$ nodes in the results as shown in Fig.11. Node $P^{i+1}$ is shifted from node $P^i$ by $(\varDelta x, \varDelta y, \varDelta z)$. The shifts are integer vectors and range from 0 to $X$, 0 to $Y$ and 0 to $Z$ along $X$ axis, $Y$ axis and $Z$ axis, respectively.

It is infeasible to exhaustively search all $(X \times Y \times Z)^{(N-1)}$ paths to find the global optimization. The reduction of the search space is based on the observation that the least cost path is in the principal direction of the candidate points. Thus, we can search the optimal path along the principal direction progressively.

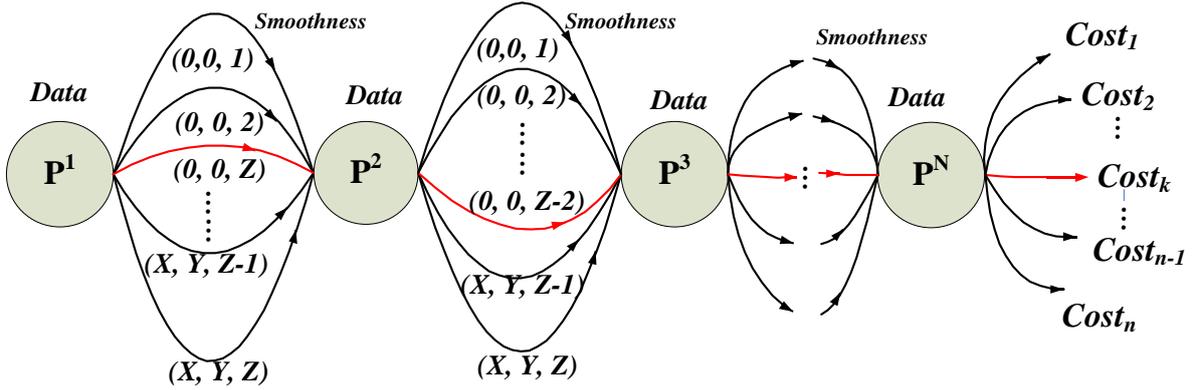

Fig.11. Graph to be optimized by the least cost path model. This graph consists of $N$ nodes, namely $P^1$, $P^2$, $P^3$,..., $P^N$. The cost of each node and edge are calculated by the data term and smoothness term. There will be $(X \times Y \times Z) \times (N-1)$ edges in the graph leading to $(X \times Y \times Z)^{(N-1)}$ paths from the node $P^1$ to the end $P^N$. Our objective is to find the least cost path among them.

Since the road is often continuous, the principal direction can be a piecewise constant in the search space. We estimate the principal direction of the path using the singular value decomposition (SVD) method. Assuming that there are $q$ candidate points in the search area, from SVD we have

$$\mathbf{D}_{q\times 3} = \mathbf{U}_{q\times q}\mathbf{S}_{q\times 3}\mathbf{V}^{T}_{3\times 3} \qquad (19)$$

where **D** is the input matrix decomposed into the matrices **U**, **S** and **V**. Denote the first, second and third column of **V** as $\mathbf{V_1}$, $\mathbf{V_2}$ and $\mathbf{V_3}$, respectively. The principal component $\mathbf{V_1}$, which corresponds to the largest eigenvalue, is chosen as the principal direction. The obtained principal direction is shown in Fig.12. The search space is fixed as $100\times100\times100$ voxel$^3$.

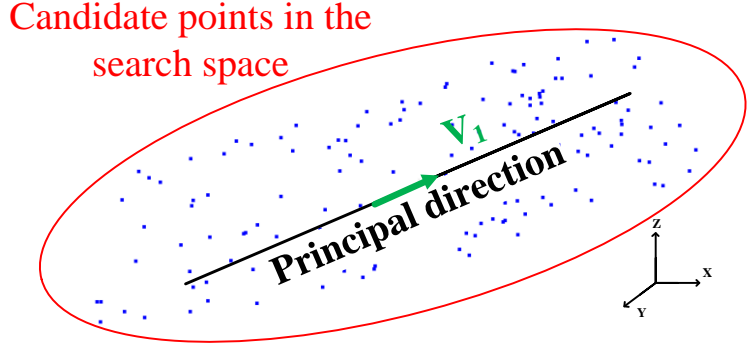

Fig.12. The principal direction obtained by SVD.

We obtain the step size by

$$(\Delta x, \Delta y, \Delta z) = \left[\left(1 - \frac{\|\mathbf{V_1}\|^2}{\sqrt{\|\mathbf{V_1}\|^2 + \|\mathbf{V_2}\|^2 + \|\mathbf{V_3}\|^2}}\right)\cdot (X_C, Y_C, Z_C)\right], \qquad (20)$$

where $X_c$, $Y_c$ and $Z_c$ are the length of $X$ axis, $Y$ axis and $Z$ axis in the current search space.

Next we propose the least cost path model (LCPM) to find the optimal path. The path proceeds along the principal direction of the search space as shown in Fig.13, where the principal direction is supposed to be the $X$ axis.

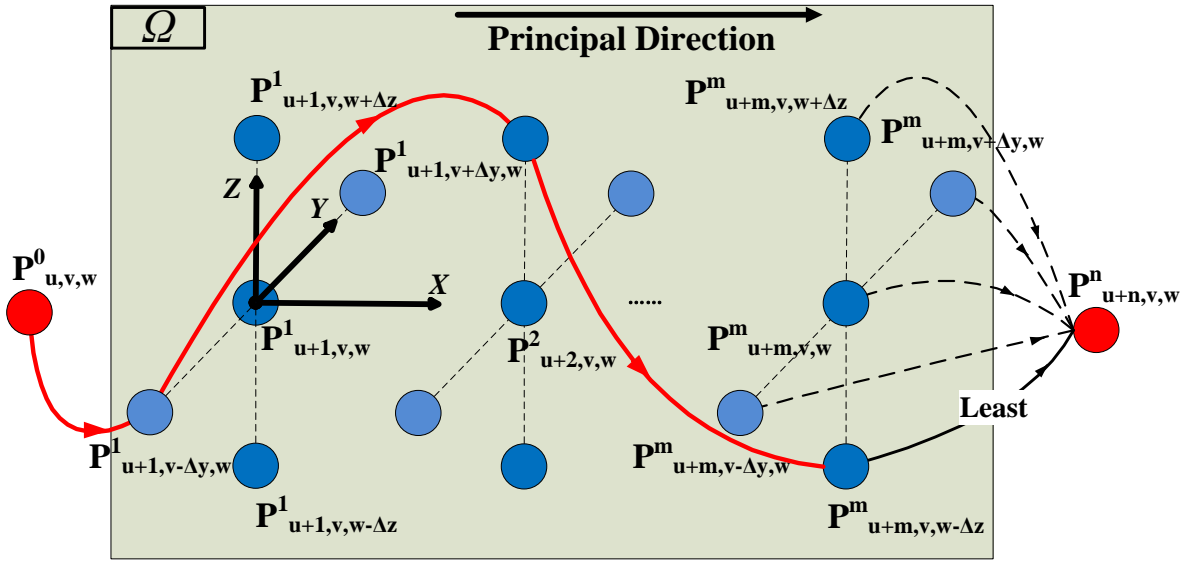

Fig.13. The new least cost path of graph $\Omega$ when added a new node $P^n_{u+n,v,w}$.

We add a starting node $P^0_{u,v,w}$ for each path to be refined. The path is from $u$ to $u+m$ in the principal direction. The cost of the connection between the starting node $P^0_{u,v,w}$ and other nodes is 0. Assuming that we have found the optimal path from the starting node to each node, now add a new node $P^n_{u+n,v,w}$ in graph $\Omega$. The least cost from $P^n_{u+n,v,w}$ to its prior node is calculated by Eq.(21). For the newly added node $i$, the optimization from the starting node to $i$ contains the solution from the starting node to its prior node $j$. $L_j$ is known, so the computation of $L_i$ incurs quite a low complexity.

$$L_i = \min_{j=1}^{num} (L_j + Data_i(u,v,w) + Smoothness_{i,j}(u,v,w)) \tag{21}$$

We store the least cost path from the starting node to each node. Each path has a cost obtained by Eq.(17). We refer to the least cost path to backtrack the complete path from the end node. Nodes in this least cost path are mapped to voxels and used to refine the incomplete curb edges. Since LCPM takes the connection costs into consideration, it can bring back the non-candidate points to obtain the optimization under the given cost function.

## 4. Experiment and results

### 4.1. Data collection

There are three main components in the mobile LiDAR scanning system, namely laser scanner, global navigation satellite system (GNSS) and inertial measurement unit (IMU) as shown in Fig.14. The scanner measures the distance between the system and the object and the angle of the launch is known to calculate the object position. The GNSS locates the global position of the scanner. The IMU is to estimate the position of the scanner when GNSS does not work.

Our mobile LiDAR data is acquired by the Riegl VMX-450 system. This laser scanner uses a narrow infrared laser beam at a very high scanning rate which can be up to 200 lines/sec and enables full 360-degree beam deflection without any gaps. Our data is collected in a 246,142.05 $m^2$ residential area consisting of various types of roads. The data size is larger than 16.7GB in 'txt' format and contains about 300 million points. The geographic location is from (51°° 4'15.12" N, 114°° 5'1.37" W) to (51°° 4'17.12" N, 114°° 4'7.47" W). The length is 1166.55 meters, the width is 211 meters and the elevation difference is 38.59 meters.

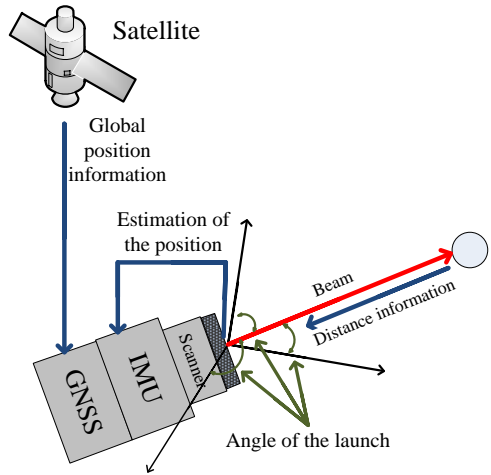

Fig.14. The process of collecting data.

Fig.15 (a) is a piece of the input data. In the pre-processing steps, we remove the non-ground areas, such as trees, houses and cars. As shown in Fig.15 (b), since the road points are much denser than non-ground areas, there is a peak in the elevation histogram. The function f′(x) is the derivative of the function f(x), which is used to describe the elevation $x$ and number of points $y$. The global extremal point in f(x) meets f′(x+ε)×f′(x-ε)<0, where $\varepsilon$ is a small positive number. When $x$ is equal to $m$, the number of points $y$ achieves the maximum. Two local extremal points in f′(x) near the point x=m are x=A and x=B. In our algorithm, the elevation from $m$-[2×($m$-$A$)] to $m$-[2×($m$-$B$)] is chosen as the ground areas as shown in Fig.15 (b). The result of removing non-ground areas is shown in Fig.15 (c).

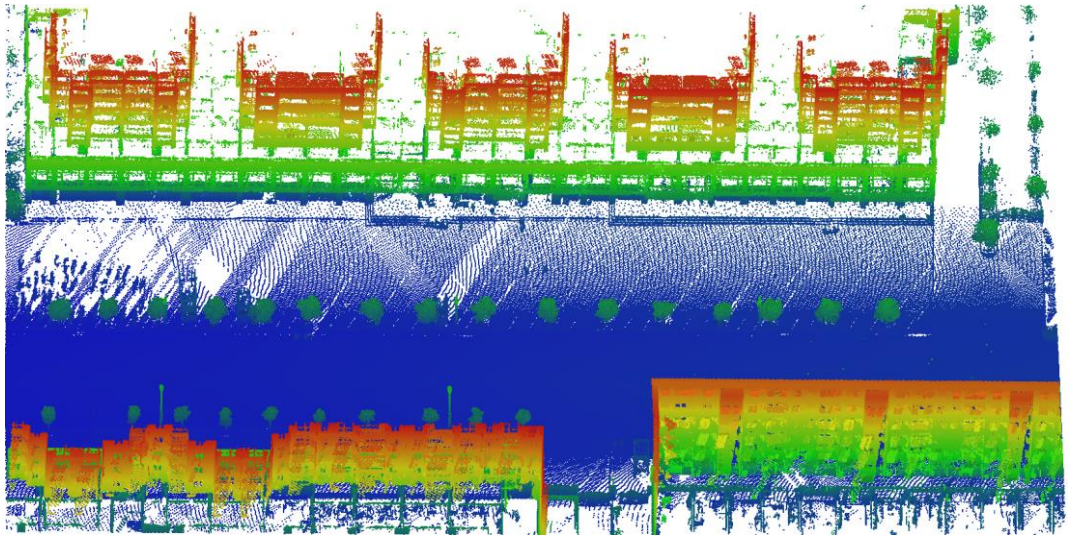

(a)

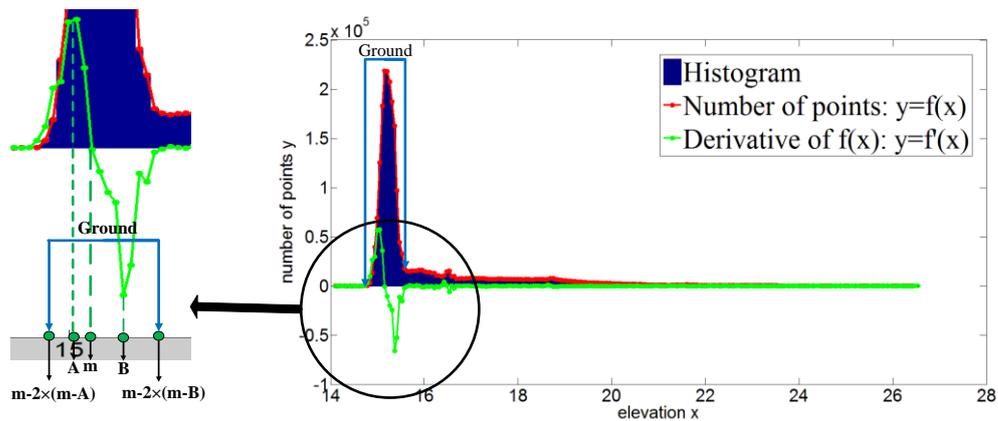

(b)

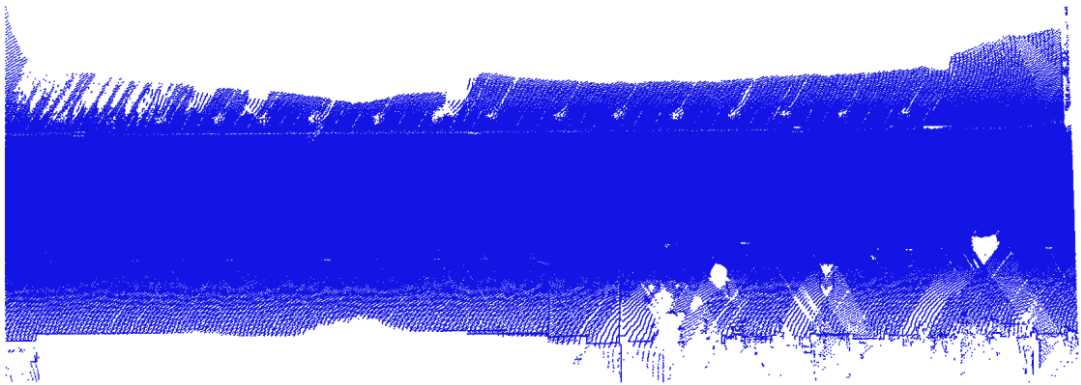

(c)

Fig.15. The process of removing non-ground regions. (a) Input point clouds. (b) The elevation histogram. (c)

The result of removing non-ground areas.

We use the voxel based representation to organize the ground point sets. The volume of each voxel is 0.04×0.04×0.04 m$^3$. The intensity of each voxel is used to calculate the sampling density gradient. Both *penaltyD* and *penaltyS* depend on the result of the curb point extraction. As shown in Fig.16, the horizontal axis means the percentage of candidate points in the search space, which is introduced in Section 3.5 and the vertical axis means the penalty. When the percentage is smaller than 0.04, we think there are no curbs in the current search space. If the extracted candidate points are limited, the penalty *penaltyD* should be small in order to consider more non-candidate points whereas the penalty *penaltyS* should be large to keep the principal direction. If they are sufficient, *penaltyD* should be large enough to consider more candidate points whereas *penaltyS* should be small in case of the zigzag curb. The penalty *penaltyV* should be large which is 1000 in our algorithm.

The process of LCPM is illustrated in Fig.17 and the optimization starts from the left to right. At the intersections, where the principal direction changes greatly, we use a curve for the missing curb areas. For occlusions, we find the optimal path based on the virtual nodes.

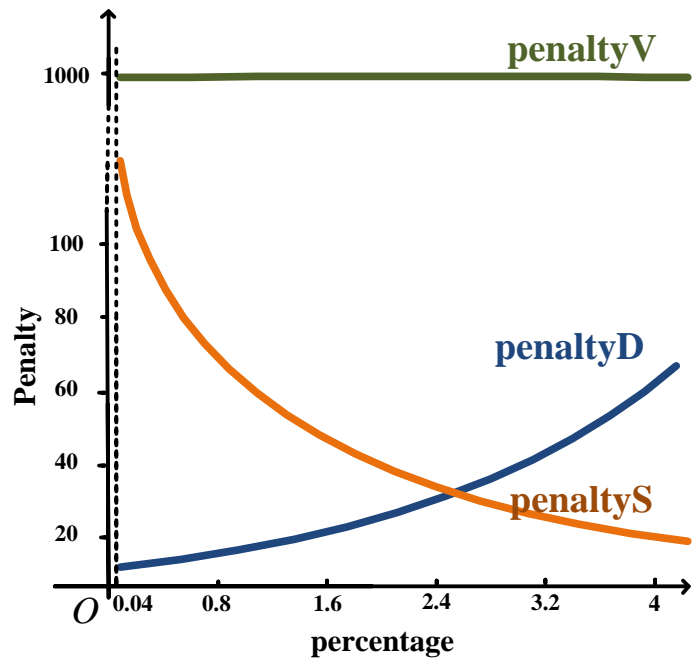

Fig.16. Selection of the penalty.

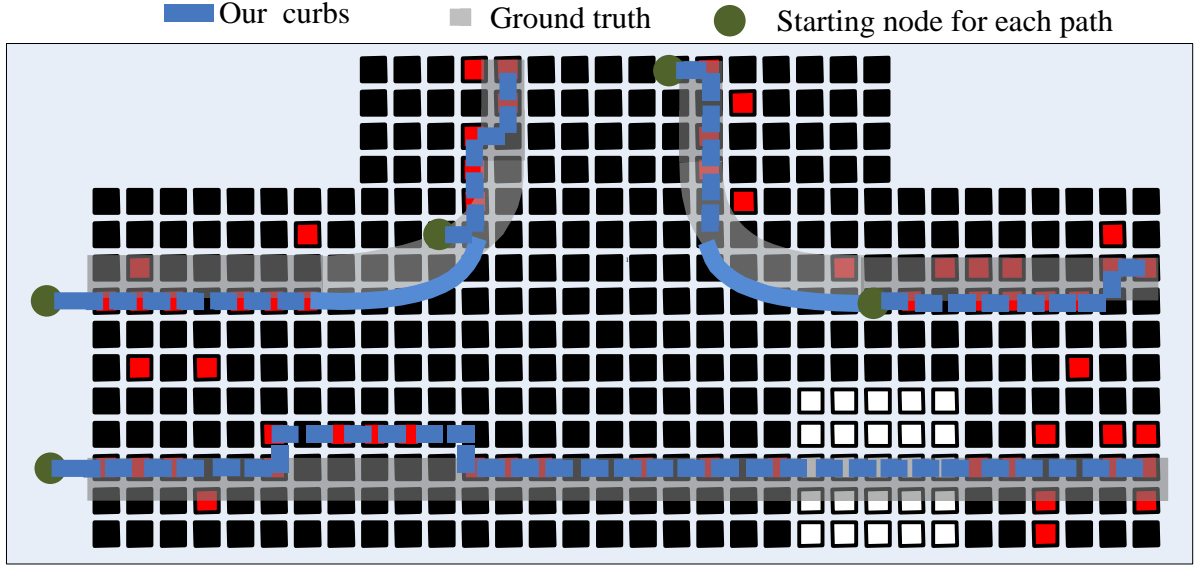

Fig.17. The process of LCPM

## 4.2. Experiments

We evaluate our algorithm in terms of three aspects: robustness, accuracy and efficiency. To further evaluate the robustness of our algorithm, we test it on two large-scale road environments, including a residential area collected by Riegl VMX-450 system (16.7GB, 300 million points) and an urban area collected by the OptechLynx scanner system (1.07GB, 20 million points).

### 4.2.1 Extraction of the curbs

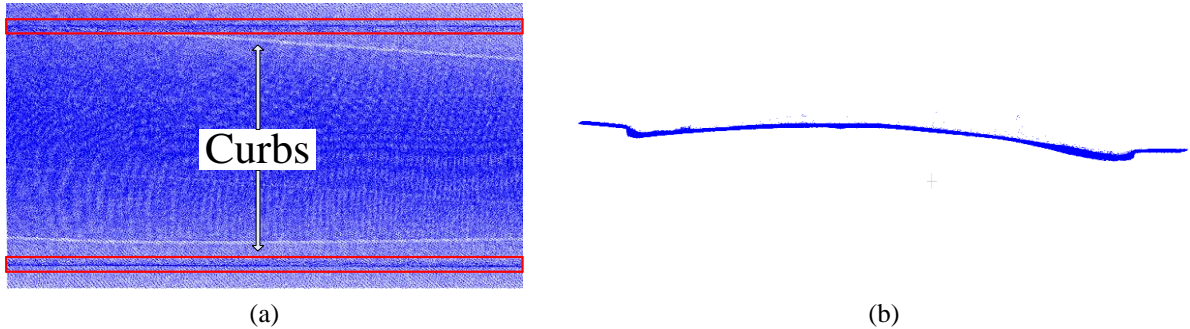

(a)                                   (b)

Fig.18. Descriptions of curbs. (a) Projection on *XOY*. (b) Projection on *XOZ*.

Fig.18 (a) and Fig.18 (b) show the projection of curbs on *XOY* and *XOZ* planes, respectively. Most of the existing methods use these projections as the input to extract curbs. As mentioned before, these methods lose all 3D information and can hardly deal with occlusions. Our algorithm uses the full 3D information of the point clouds and can deal with the challenging situations caused by either scanner system or complex road environments.

The following are the results of our method on different challenging situations. To better show the results, we highlight results by large red points.

(1) Uneven density

The curb points may be sparse or dense as shown in Fig.19 (a) and 19 (d). The extracted candidate points are shown in Fig.19 (b) and Fig.19 (e), respectively. The candidate curbs are noisy and incomplete. By using LCPM, we link them into the optimal curbs as shown in Fig.19 (c) and Fig.19 (f), respectively.

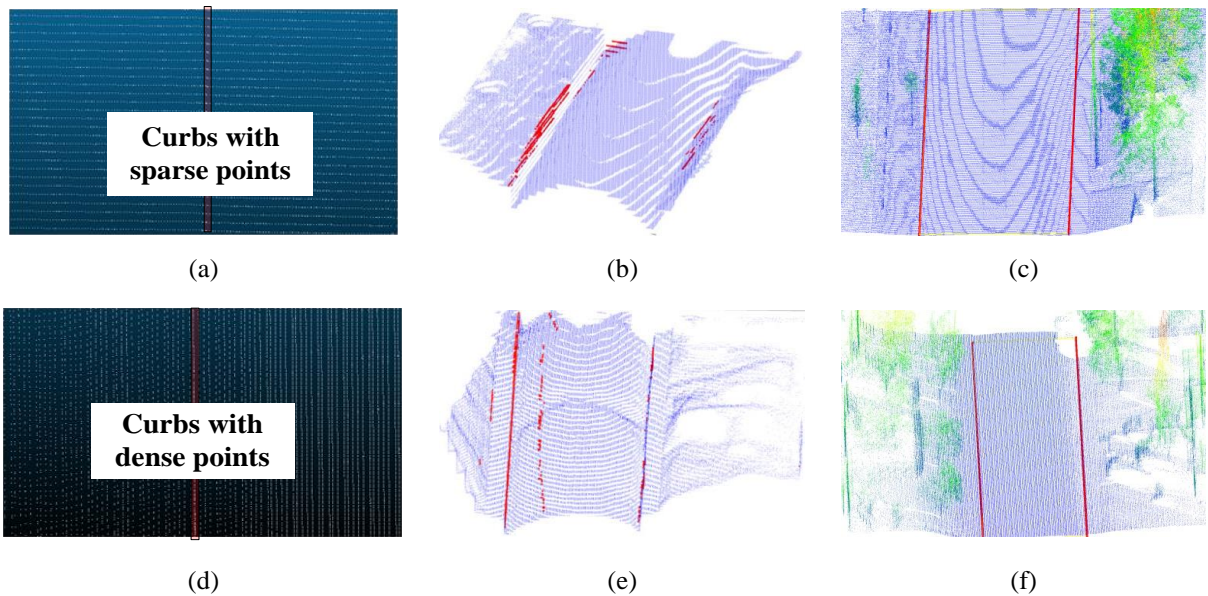

Fig.19. Uneven curb points. (a) Curbs with sparse points. (b) Extracted candidate points of (a). (c) The refinement of (b). (d) Curbs with dense points. (e) Extracted candidate points of (d). (f) The refinement of (e).

An example of varying densities between right and left side of the road is shown in Fig.20. The various densities lead to the unreliable computation of our sampling density gradients. This causes undesirable extraction as shown in Fig.20 (b). As seen from Fig.20 (c), although the various densities may cause incorrect candidate points extracted, our model still can obtain the desirable curbs.

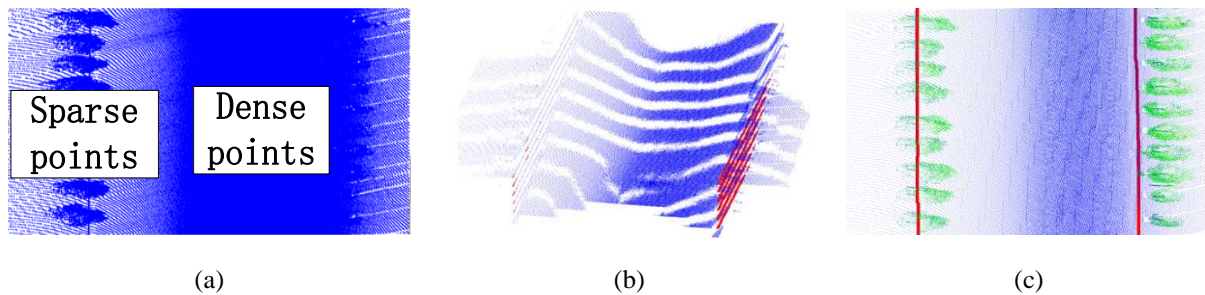

Fig.20. Various densities. (a) The densities of points are varying from left side to right side of the road. (b) Extracted candidate points of (a). (c) The refinement of (b).

The density of the point clouds collected from various systems is different. To test our algorithm, we down-sample the road point clouds to different cases as shown in Fig.21 (a)-(d). Results show that our method is robust to the sparsity. Even for the case where the point clouds are downsampled to 1%, the proposed method can still extract the curbs, which is difficult for any existing methods.

One challenging problem as shown in Fig.21 (a) is that there are gaps in point clouds caused by the MLS itself. These gaps are easy to be wrongly detected as curbs but well filtered by our method.

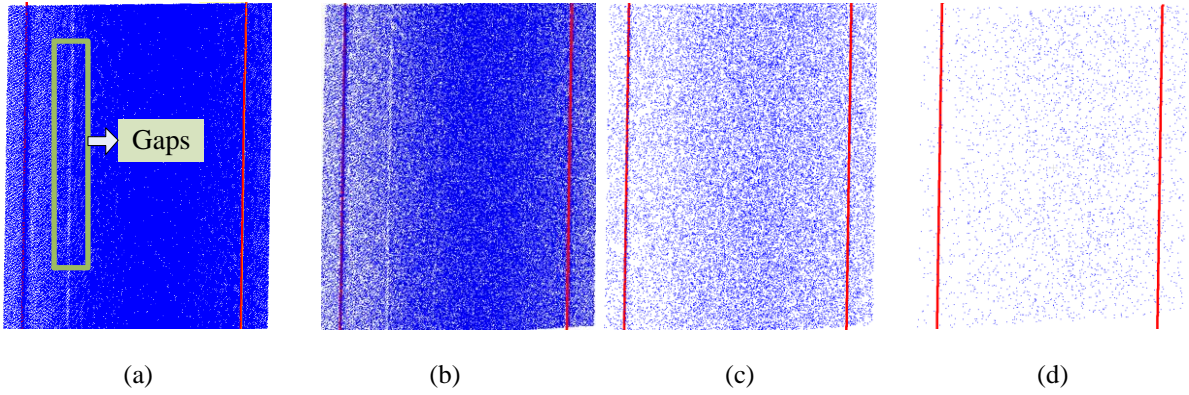

(a)              (b)              (c)              (d)

Fig.21. Sparse points. Sparse road by sampling data in (a) 100%, (b) 50%, (c) 10% and (d) 1%.

(2) Missing curb areas

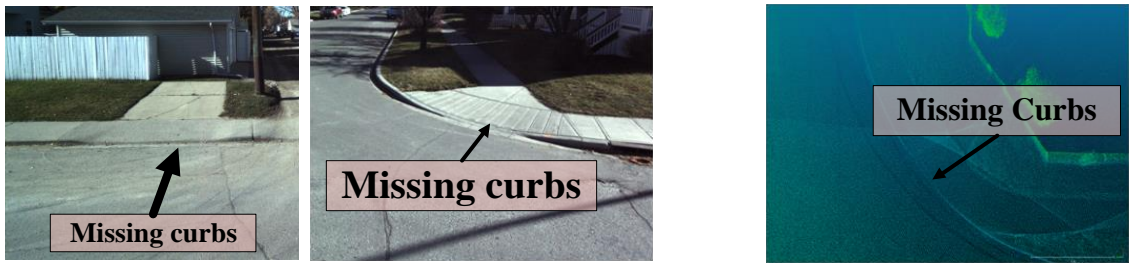

(a)                                     (b)

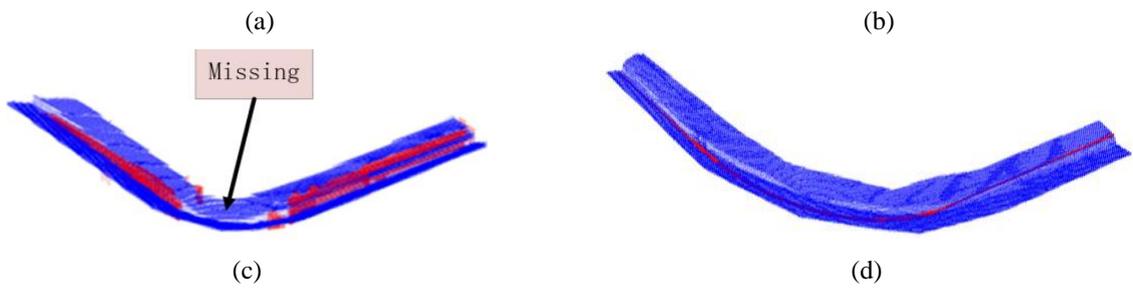

(c)                                     (d)

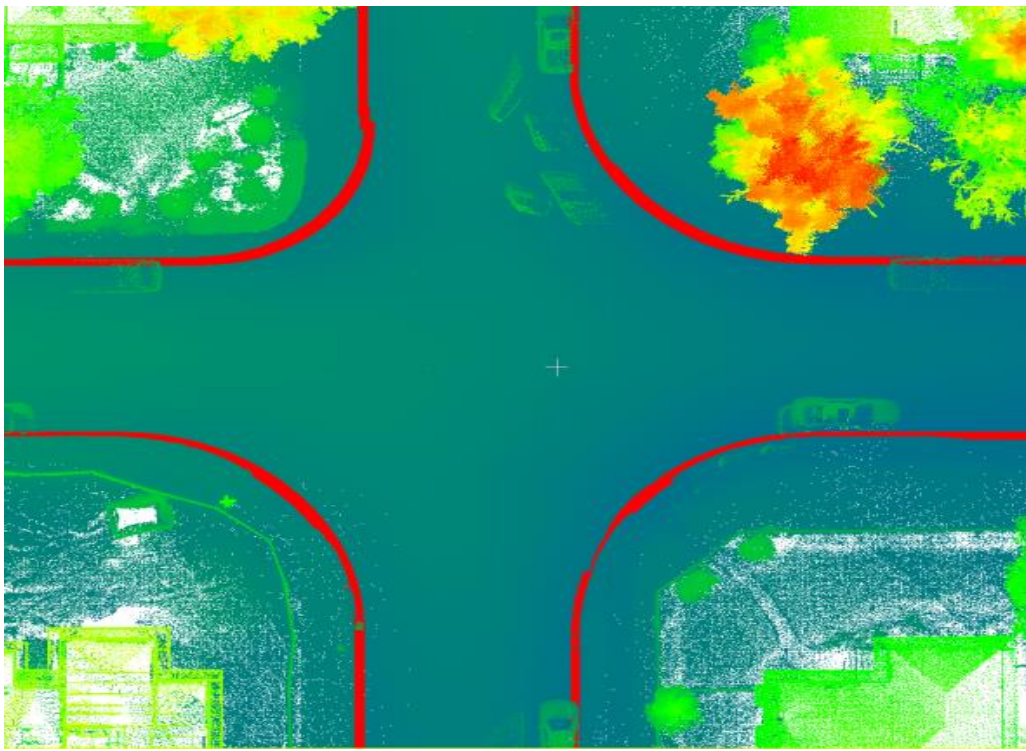

(e)

Fig.22. Missing curbs. (a) Missing curb areas in 2D image. (b) Missing curb areas in 3D point clouds. (c) Missing candidate points in curb areas. (d) Our refined result of (c). (e) Results of an entire crossroad.

The curb on the road may be missing as shown in Fig.22 (a), which is designed for wheelchairs and bicycles. There is no curb information in these areas as shown in Fig.22 (b). This is our limitation, because there is only one large sampling density gradient in these areas. The candidate points are missing totally as shown in Fig.22 (c). If missing curb areas are along the straight road, we can obtain the complete curbs based on the neighbor information. However, if these areas are at the intersection road, we can only use a fixed curve based on the prior knowledge to link curbs as shown in Fig.22 (d). Fig.22 (e) is the result of an entity crossroad to demonstrate our results.

(3)   Slope and occlusion

To test the sloping road, we lift one side of the road up to 30 degrees as shown in Fig.23 (a). This is difficult for extraction algorithms based on the elevation. For occlusions caused by cars or pedestrians on the road as shown in Fig.23 (b), LCPM considers the virtual nodes to find the optimal path.

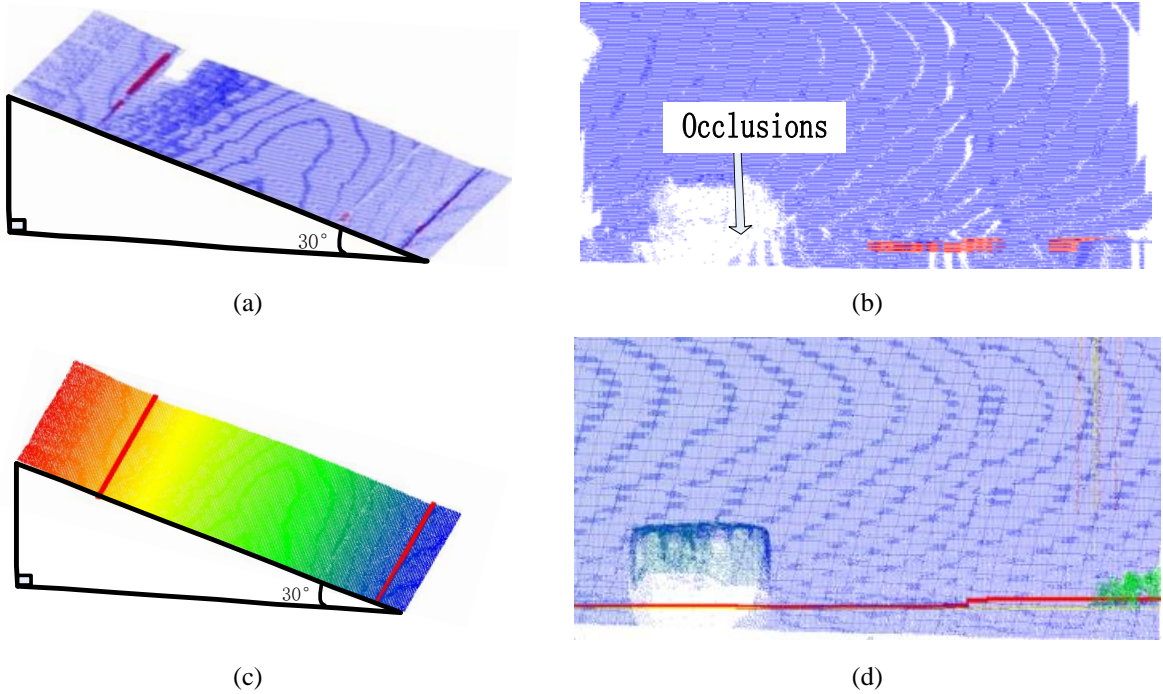

Fig.23. Slope and occlusion. (a) Sloping road. (b) Occluded road. (c) Results of (a). (d) Results of (b)

The change $C$, which is used to calculate the energy, is independent of the coordinate system. Thus, our method is invariant to the rotation and hence, curbs in the slope are well extracted with our algorithm. Extracted candidate points and refinements are shown in Fig.23 (a) and 23 (c).

As opposed to the missing curb areas, there are no points in occlusions. In our algorithm, occlusions are filled with virtual nodes. To connect the virtual nodes, a large *pentaltyV* is used. For a small occlusion, the optimal path passes through the missing curb areas as shown in Fig.23 (d). However, for a large occlusion, we empirically conclude that there is no curb when the percentage of candidate points is lower than 0.04. Extracted candidate points and refinements are shown in Fig.23 (b) and 23 (d).

(4)   Large-scale experiments

The residential area mostly contains trees, parking cars and houses. Fig.24 (a)-24 (f) correspond to six parts of the residential area. As shown in Fig.24 (a), we zoom into three areas to show the results, including the straight curbs A, the intersection area B and the occluded area C. Curbs in these areas are well extracted and refined.

The urban area mostly contains trees, buildings and traffic facilities as shown in Fig.24 (g). We zoom into two areas to show the results, including the occluded area A and an alley B. In both the areas, curbs are well extracted.

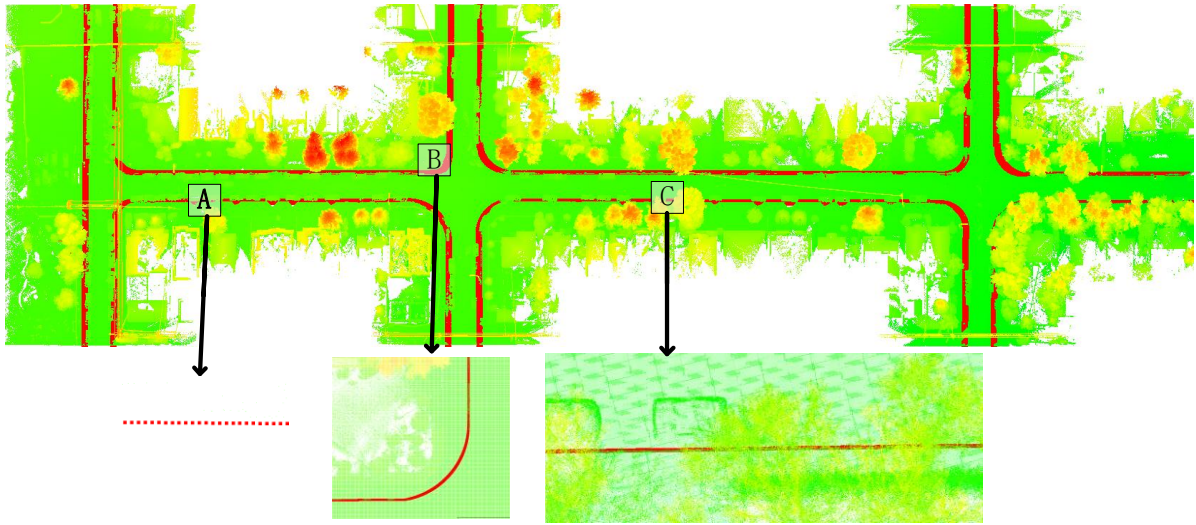

(a)

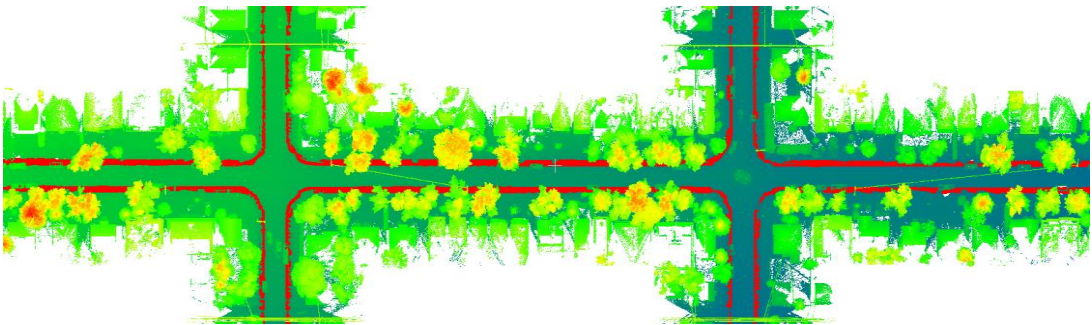

(b)

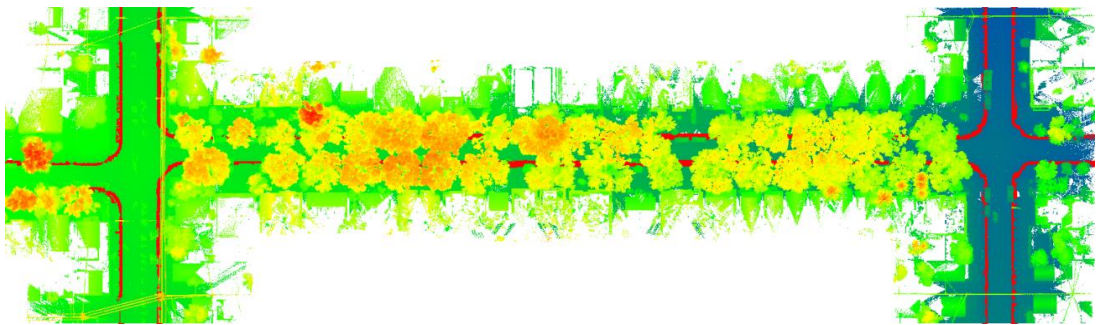

(c)

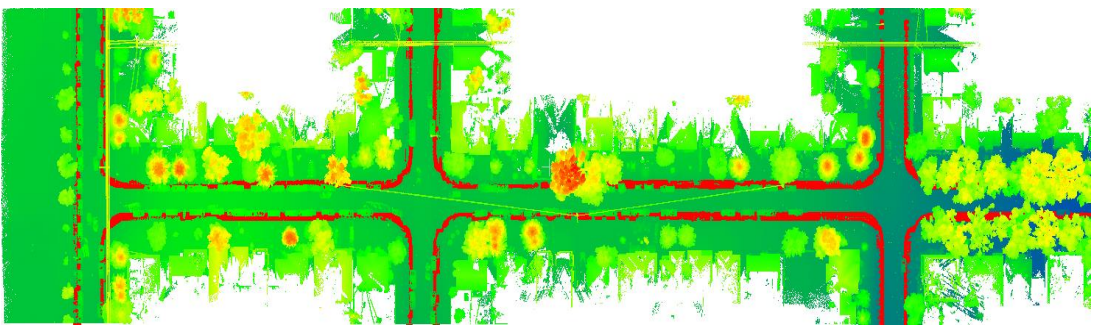
19

(d)

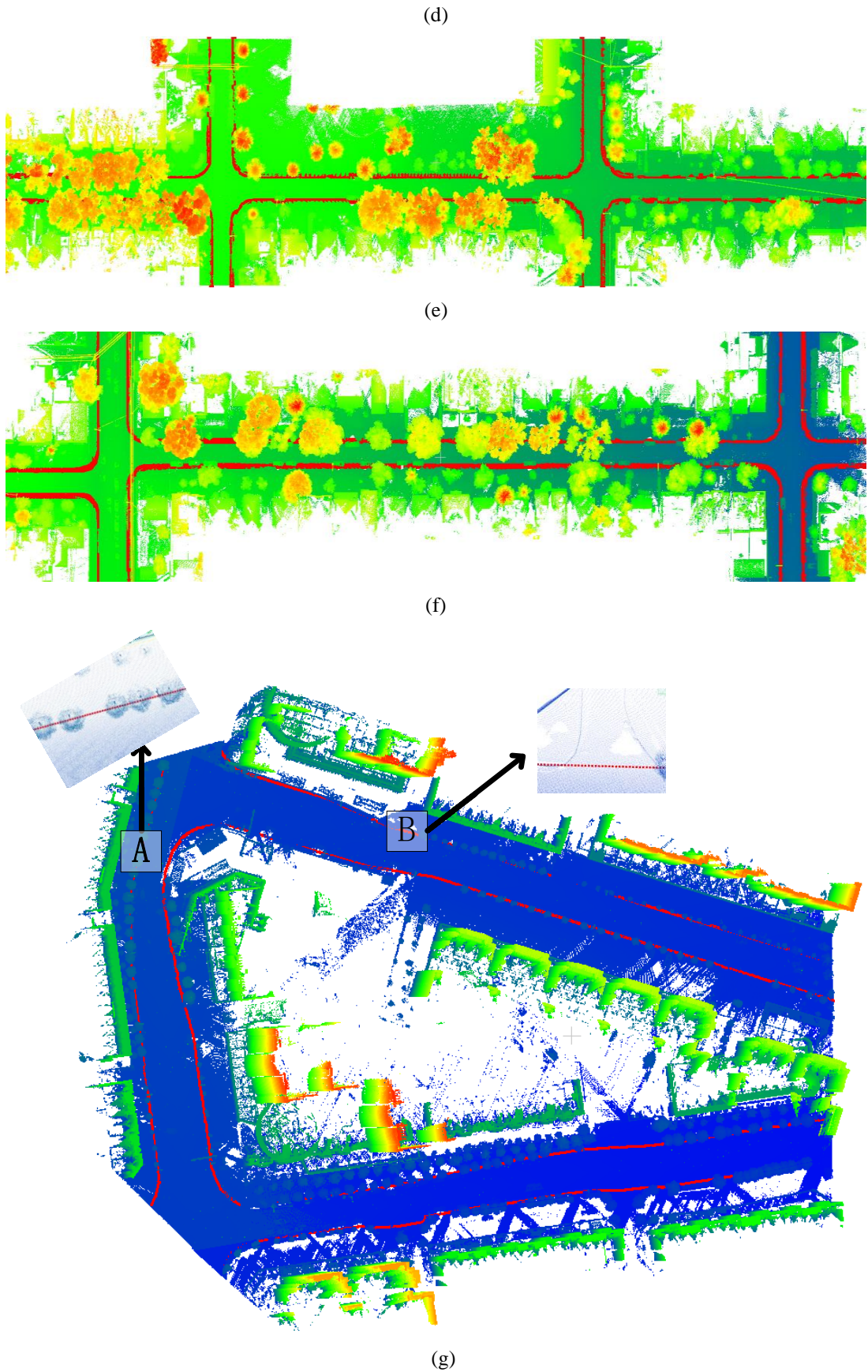

(e)

(f)

(g)

Fig.24. Experimental results, including a large residential area and an urban area. (a) Results of Part 1 from the large-scale residential area. (b) Results of Part 2 from the large-scale residential area. (c) Results of Part 3 from the large-scale residential area. (d) Results of Part 4 from the large-scale residential area. (e) Results of Part 5 from the large-scale residential area. (f) Results of Part 6 from the large-scale residential area. (g) Results of the large-scale urban area.

To visualize the difference between our results and the ground truth, we show the candidate points, resultant curbs, ground truth and the overlap in Fig.25 refers to the above Part 1.

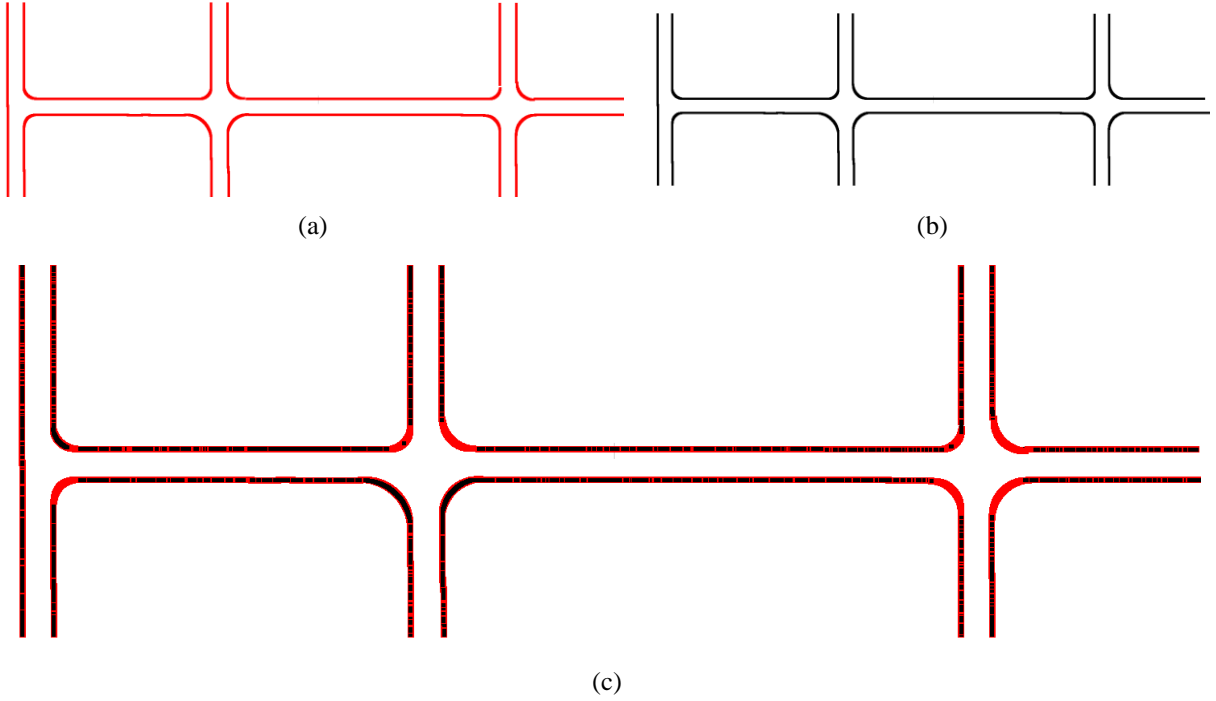

Fig.25. The visualization of the difference between our results and the ground truth for Part 1. (a) Our results. (b) Ground truth. (c) Overlap between (a) and (b).

Results in Fig.24 and Fig.25 demonstrate that our method is robust against large-scale data testing. We also use videos to show our results in 3D in supplementary materials.

*4.2.2  Quantitative evaluation*

In this section, we quantify the difference between our results and the ground truth and compare our method with other related work.

Assuming there is a point $p$ in the clouds, $L$ is the curb obtained by our method and $L'$ is the ground truth obtained by the manual method. For the classification, there are four results of $p$, namely TP (true positive) if $p \in L \bigcap L'$, TN (true negative) if $p \in \overline{L} \bigcap \overline{L'}$, FP (false positive) if $p \in L \bigcap \overline{L'}$ and FN (false negative) if $p \in \overline{L} \bigcap L'$ as shown in Fig.26. We evaluate the difference in terms of 4 aspects based on TP, TN, FP and FN [26], namely true positive rate(TPR), true negative rate(TNR), positive predictive value(PPV), negative and predictive value(NPV).

We need a parameter $D$ to decide whether the test point $p$ belongs to $L$ or $L'$. If the distance between $p$ and $L$ or $L'$ is less than $D$, $p \in L$ or $p \in L'$, else $p \in \overline{L}$ or $p \in \overline{L'}$ as shown in Fig.26. In the following, we show the accuracy under different $D$ in TABLE I.

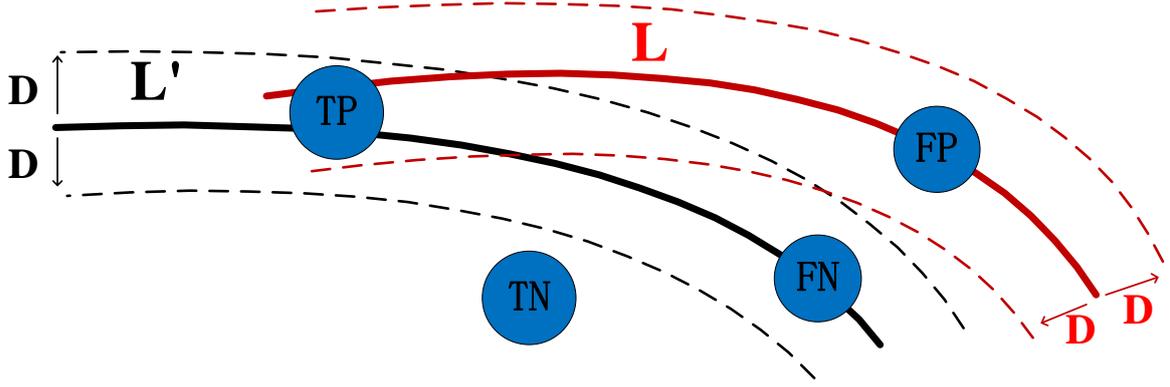

Fig.26. Four types of results for each point, namely TP, TN, FP and FN. *L* is the result of our method and *L'* is the ground truth. The width of *L* or *L'* is 2×*D*.

TABLE I reports the quantitative results of the above large-scale experiments, where *SL*, *Int* and *All* respectively denote the straight areas, intersections and all areas. The extracted curbs consist of 294,418 points and 115,971 of them belong to straight areas and others are intersections. The ground truth contains 115,240 points and 24,258 of them belong to straight areas and others are intersections. TPR is TP/(TP+FN) means the probability of true curbs that can be extracted, which is the completeness of curbs. From TABLE I, it can be observed that when the distance parameter *D* is 0.4 meters, the completeness of curbs is up to 78.62%. PPV is TP/(TP+FP) means that the extracted curbs belong to the true curbs, which is the correctness of the curbs. When *D* is 0.4 meters, the correctness of curbs is up to 83.29%. TNR is TN/(FP+TN) means the probability of non-curb areas that can be extracted, which is the completeness of non-curbs. NPV is TN/(TN+FN) means the extracted non-curb points belong to the non-curb areas, which is the correctness of the non-curbs. TABLE I indicates that at intersections, our method has a poor performance, mainly due to the absence of curb information.

TABLE I
QUANTITATIVE EVALUATION

| Evaluation (%) | | D (m) | | | | |
|---|---|---|---|---|---|---|
| | | 0.4 | 0.2 | 0.12 | 0.08 | 0.04 |
| **TPR** | SL | 89.49 | 81.30 | 66.55 | 59.13 | 26.71 |
| | Int | 61.87 | 41.21 | 26.06 | 18.66 | 8.09 |
| | All | **78.62** | **66.25** | **51.37** | **43.63** | **19.40** |
| **TNR** | SL | 99.77 | 99.70 | 99.61 | 99.56 | 99.73 |
| | Int | 99.52 | 99.64 | 99.75 | 99.82 | 99.93 |
| | All | **99.65** | **99.67** | **99.68** | **99.69** | **99.83** |
| **PPV** | SL | 91.54 | 78.93 | 56.07 | 36.01 | 10.22 |
| | Int | 69.37 | 48.84 | 32.22 | 21.26 | 7.46 |
| | All | **83.29** | **69.00** | **49.15** | **32.33** | **9.64** |
| **NPV** | SL | 99.71 | 99.74 | 99.75 | 99.83 | 99.92 |
| | Int | 99.34 | 99.51 | 99.67 | 99.79 | 99.93 |
| | All | **99.52** | **99.62** | **99.71** | **99.81** | **99.92** |

*4.2.3 Comparison with existing methods*

1) Detection methods

We compare our algorithm with related detection methods, including Yu et al. (EEC) [6], Kellner et al. (IEPF) [10], Rodríguez-Cuenca et al. (TML) [16] and Yang et al. (CS) [11], using the TPR and PPV to evaluate the completeness and correctness of curbs.

There are three steps in EEC, namely elevation gradient computation, elevation gradient filtering and curb corner point selection. This method is based on the elevation filtering, which fails when curbs are in quite different elevations or occluded. The IEPF uses iterative end-point fitting algorithm to segment the scenes. This algorithm relies on the elevation of curb and the flatness of the road to detect the curbs, which is difficult to work in complex urban scenes. TML uses the projection to detect curbs by three steps, namely thresholding, morphological processing and linear feature detection. Each step needs many parameters which are difficult to choose. CS is proposed to detect road curbs by pre-defined curb model, which is based on the elevation jump, point density and slope change.

We compare with the above mentioned methods using the data of Part 1. As shown in Fig.27, our method is much better than the above mentioned methods in detecting curbs, especially when $D$ is small. The accuracy of TML can be high, but it relies on the parameters heavily, which is difficult to tune.

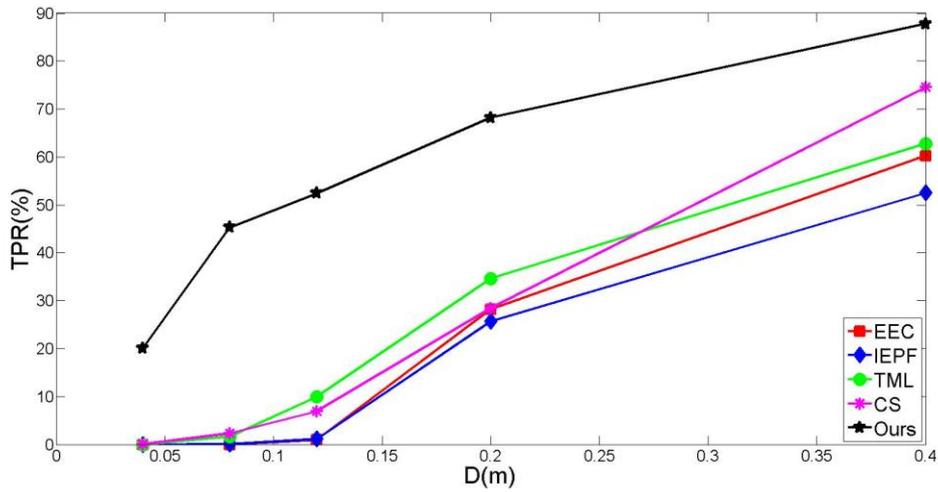

(a)

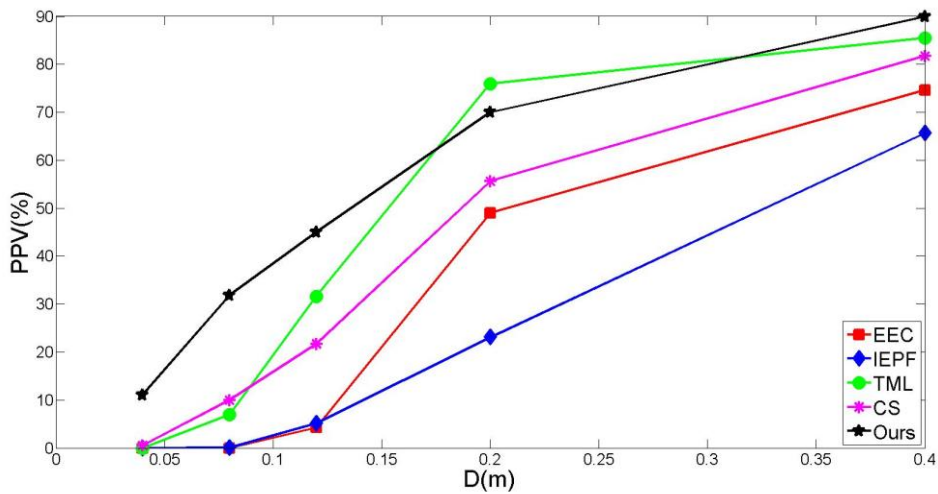

(b)

Fig.27. Comparison with the existing curb detection methods, including EEC, IEPF, TML and CS. (a) TPR to evaluate the completeness. (b) PPV to evaluate the correctness.

2) Refinement methods

We also compare our method with some typical refinement methods, including Least Square (LS), Hough Transform (HF) and Random Sample Consensus (RANSAC). We test the robustness of our method against noise by adding a random number from $-T\times d$ to $T\times d$ to the coordinate of each point, where $d$ is the minimum point distance between two points and $T$ is to set the range.

For a small random noise ($T=2$), the candidate points extracted by our method are shown in Fig.28 (a). Increasing the level of the random noise ($T=4$), we achieve an undesirable result as shown in Fig.28 (b). These candidate points, containing outliers, are the input for the later refinement.

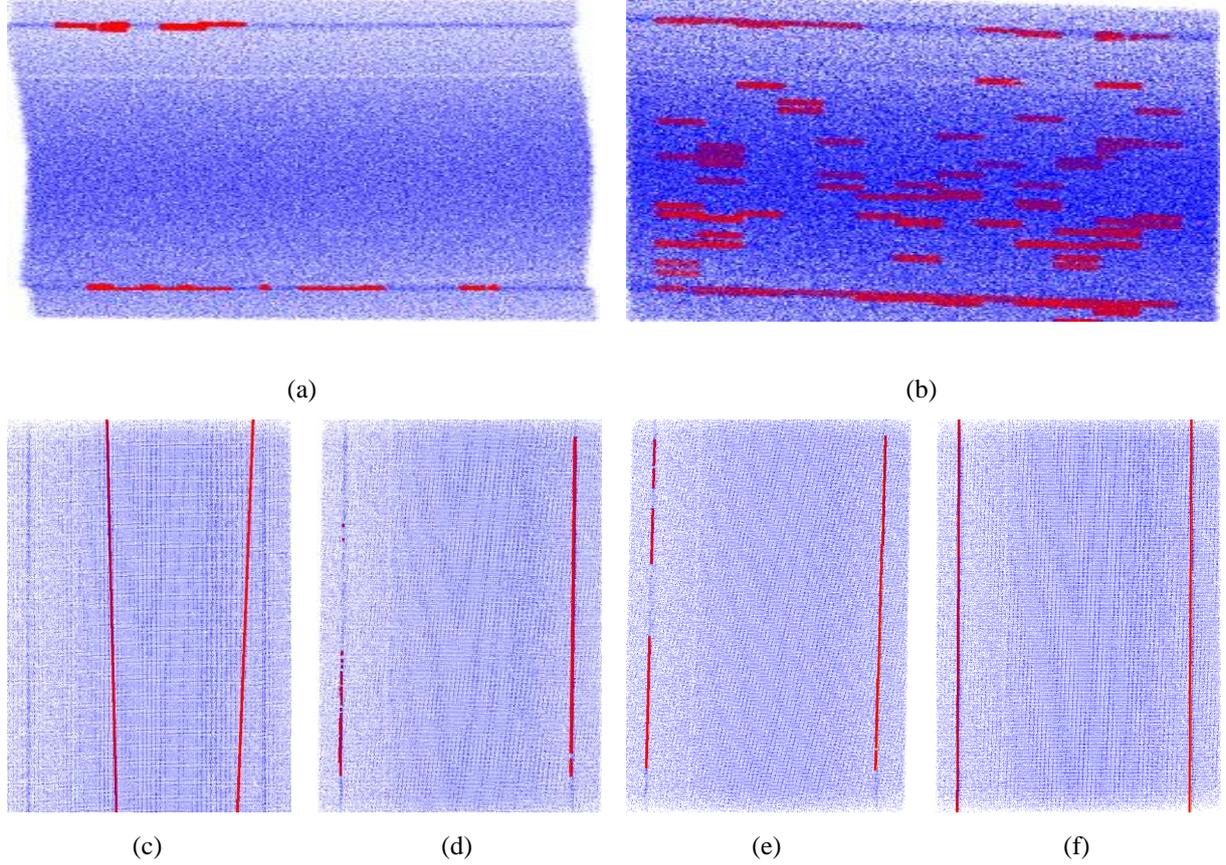

Fig.28. Comparison with the existing refinement methods. (a) Extracted candidate points from a small random noise ($T=2$). (b) Extracted candidate points from a large random noise ($T=4$). (c) Refinement by LS. (d) Refinement by HT. (e) Refinement by RANSAC. (f) Refinement by LCPM.

The results of LS, HT, RANSAC and our method are shown from Fig.28 (c)-28 (e), respectively. LS fails to refine curbs due to incorrect mean point by outliers and HT and RANSAC obtain acceptable curbs from our candidate points. 3D HT obtains the curb correctly while requires the best line in the exactly the same plane which is not always the case in point clouds. For RANSAC, only if all inliers are obtained, the line can be fitted accurately. Moreover, all these typical refinement methods do not consider the cost to connect each point. If there are few candidate points or few straight lines, both HT and RANSAC can hardly extract curbs. In this condition, LCPM obtains the optimal curbs by considering both candidate and non-candidate points which enhances the robustness considerably as shown in Fig.28 (f).

*4.2.4  Computational complexity*

Normally, there are three steps in each algorithm, namely the generation of the region of interest (ROI), the extraction of candidate curb points and the refinement of incomplete curbs. The complexity of the extraction in each algorithm is described in the following.

EEC relies on the curb profiles vertical to the road surface and 10-20 cm above the road to generate ROI. The generation uses a thresholding method ($O(N)$). This is followed by the extraction of the candidate points, including elevation gradient computation ($O(N)$), elevation gradient filtering ($O(N)$) and curb points selection ($O(N)$). Finally, the refinement as mentioned in their paper can be LS ($O(N^3)$).

IEPF calculates the distance between each point and a straight line, obtained by connecting the start and end point of a scan line, to generate the lateral distance ($O(N)$) in the preprocessing. The points are segmented by the lateral distance and classified by the clustering algorithms [27] ($O(N^2)$). Then the authors use the decision tree ($O(R \times N)$, where $R$ is the depth of the tree) based on three properties, namely mean height ($O(N)$), angle ($O(N)$) and variance ($O(N)$), to obtain the curbs. The refinement is based on the trajectory of the car. The similar candidate curbs, which show similarity in the lateral distance, are connected in linear time ($O(N)$).

TML uses elevation thresholding to obtain the ROI ($O(N)$). The maximum and minimum thresholds are set to avoid extraction of points from the road. Then the morphological processing methods are used to obtain the candidate curbs (no less than $O(2 \times N^2)$), including erosion to remove the isolated points and dilation to increase the curb candidate set of points. Refinement is based on the line feature, which is calculated by thresholding the percentage of rows or columns higher than a preset percentage, to connect the candidate points ($O(N)$). Then the authors perform a rotation around the axis to determine the edges based on the trajectory of the vehicle ($O(N)$).

CS groups point clouds into road cross sections ($O(N)$) based on GPS time in the preprocessing step. Then they use a sliding window, to extract candidate road areas based on the fact that road points at one cross section have the identical elevation ($O(N)$), to obtain ROI. The authors detect curbs based on their three proposed models, including elevation jump ($O(N)$), point density ($O(N^2)$) and slope change ($O(N)$). The refinement of the candidate curbs includes using K-nearest neighbor to cluster them ($O(N^2)$), removing fake curbs that contain few points ($O(N)$) and connecting the curbs that are sorted along the direction of the curbs ($O(N)$).

Our method calculates the histogram of the elevation to generate ROI ($O(N)$) and then calculate sampling density gradients in each axis direction to obtain the energy for each point ($O(3 \times N)$) followed by the refinement LCPM ($O(N^3)$).

We show all the computational complexity in TABLE II. From this table, the complexity of ours is the same as existing methods in the preprocessing step and much lower than IEPF, TML and CS in the extraction. EEC has a low complexity, because it only depends on the unreliable elevation difference. For the refinement, our complexity is higher, because we do not use any extra information, such as trajectory for IEPF and TML, or GPS time for CS. However, only our refinement method achieves global optimization.

TABLE II
COMPLEXITY OF EACH ALGORITHM

| Method | Preprocessing | Extraction | Refinement |
|---|---|---|---|
| EEC | $O(N)$ | $O(N)$ | $O(N^3)$ |
| IEPF | $O(N)$ | $O(N^2)$ | $O(N)$ |
| TML | $O(N)$ | $O(N^2)$ | $O(N)$ |
| CS | $O(N)$ | $O(N^2)$ | $O(N^2)$ |
| Ours | $O(N)$ | $O(N)$ | $O(N^3)$ |

### 4.2.5 *Results on the 2D map*

We overlay our results on the images from Google Earth. As shown in Fig.29, our results can match the curbs in the map accurately.

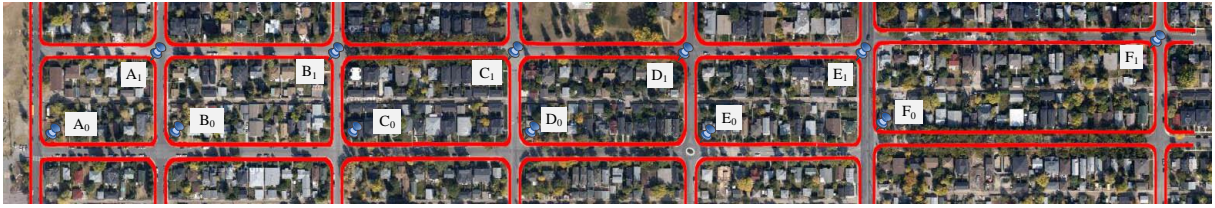

Fig.29. Overlay of our results on the Google Earth.

To quantify our results on 2D map, we compare the length (*L*) and width (*W*) of selected six places in Google Earth with our results in TABLE III.

TABLE III

LENGTH AND WIDTH OF THE SELECTED SIX AREAS

| Map (*m*) | | Ours (*m*) | | Geographic location |
|---|---|---|---|---|
| *L* | *W* | *L* | *W* | |
| 108 | 87 | 110.93 | 86.74 | $A_0$(51.071492N,-114.08303W), $A_1$(51.072267N,-114.08152W) |
| 157 | 86 | 158.57 | 85.58 | $B_0$(51.071506N,-114.08128W), $B_1$(51.072267N,-114.07904W) |
| 158 | 88 | 159.94 | 85.67 | $C_0$(51.071516N,-114.07884W), $C_1$(51.072276N,-114.07657W) |
| 157 | 85 | 158.69 | 86.10 | $D_0$(51.071498N,-114.07637W), $D_1$(51.072269N,-114.07414W) |
| 157 | 84 | 157.33 | 83.87 | $E_0$(51.071510N,-114.07392W), $E_1$(51.072262N,-114.07167W) |
| 268 | 87 | 270.65 | 87.61 | $F_0$(51.071584N,-114.07143W), $F_1$(51.072362N,-114.06756W) |

From TABLE III, the difference between our results from the above detected curbs and the Google Earth is 0.006 per meters and the mean square error is 2.67. These evaluations show that our results are accurate and reliable.

## 5. Conclusions

Curb extraction is essential for understanding road environments. This paper presents a robust, accurate and efficient solution for road curb extraction from mobile LiDAR point clouds. To the best of our knowledge, this is the most comprehensive work on road curb extraction from point clouds. We evaluate the proposed method on a large-scale residential area and an urban area. Our algorithm works effectively for large-scale mobile LiDAR point clouds. Different quantitative evaluations, including true positive rate (TPR), true negative rate (TNR), positive predictive value (PPV) and negative predictive value (NPV), indicate that our method is more accurate than existing algorithms.

Possible directions for future research include segmentation of scenes, classification of objects and understanding of traffic environments.

**Acknowledgment**

The authors would like to thank the City of Calgary and Optech Corporation for providing the mobile LiDAR data.